\documentclass{article}

\usepackage{PRIMEarxiv}
\usepackage[utf8]{inputenc} 
\usepackage{amsmath,amsfonts,amssymb}
\usepackage{mathtools}
\usepackage[hidelinks]{hyperref}

\usepackage{float}

\usepackage{subcaption}
\usepackage{array}
\usepackage{booktabs}
\usepackage{url}
\usepackage{xcolor}
\usepackage{graphicx}
\usepackage{multirow}
\usepackage{makecell}
\usepackage{caption}
\usepackage[euler]{textgreek}
\usepackage{relsize}
\usepackage{amssymb}
\usepackage{dblfloatfix}
\usepackage[subnum]{cases}
\usepackage{algorithm}
\usepackage{algpseudocode}

\usepackage{physics}

\renewcommand{\mathbf}{\boldsymbol}

\usepackage{float}
\usepackage{capt-of}
\usepackage{tabularx}
\newcount\Comments  
\usepackage{color}
\definecolor{darkgreen}{rgb}{0,0.5,0}
\definecolor{purple}{rgb}{1,0,1}
\definecolor{amber}{rgb}{1,0.49,0}
\definecolor{uglycolor}{rgb}{0.5,0.5,0}
\newcommand{\kibitz}[2]{\ifnum\Comments=0\textcolor{#1}{#2}\fi}

\makeatletter
\newcommand*{\rom}[1]{\expandafter\@slowromancap\romannumeral #1@}
\makeatother

\title{Learn2Drive: A neural network-based framework for socially compliant automated vehicle control
\thanks{\textit{\underline{Citation}}: 
\textbf{Liu et al. Learn2Drive: A neural network-based framework for socially compliant automated vehicle control.}} 
}

\author{
  Yuhui Liu \\
  Department of Civil Engineering \\
  Saint Louis University \\
  \AND
  Samannita Halder \\
  Department of Electrical and Computer Engineering\\
  Saint Louis University \\
  \AND
  Shian Wang \\
  Department of Civil, Environmental, and Architectural Engineering\\
  The University of Kansas \\
  \And
  Tianyi Li \\
  Department of Civil Engineering \\
  Saint Louis University \\
  \texttt{tianyili.ai@gmail.com} 
  }

\begin{document}
\maketitle


\begin{abstract}
This study introduces a novel control framework for adaptive cruise control (ACC) in automated driving, leveraging neural networks and physics-informed constraints. As automated vehicles (AVs) adopt advanced features like ACC, transportation systems are becoming increasingly intelligent and efficient. However, existing AV control strategies primarily focus on optimizing the performance of individual vehicles or platoons, often neglecting their interactions with human-driven vehicles (HVs) and the broader impact on traffic flow. This oversight can exacerbate congestion and reduce overall system efficiency. To address this critical research gap, we propose a neural network-based, socially compliant AV control framework that incorporates social value orientation (SVO). This framework enables AVs to account for their influence on HVs and traffic dynamics. By leveraging AVs as mobile traffic regulators, the proposed approach promotes adaptive driving behaviors that reduce congestion, improve traffic efficiency, and lower energy consumption. Numerical results demonstrate the effectiveness of the proposed method in adapting to varying traffic conditions, thereby enhancing system-wide efficiency. Specifically, when the AV's control mode shifts from prioritizing its own energy conservation to optimizing collective traffic flow efficiency, the controlled AV proactively adapts its acceleration profile. This prosocial behavior yields at least a 38.39\% improvement in the average speed of downstream vehicles and effectively dampens traffic oscillations, demonstrating significant enhancements in system-wide dynamics. 
The implementation code is available \url{https://github.com/lilab2024/Learn2Drive-SVO_v1.git}.
\end{abstract}


\section{Introduction}\label{sec:introduction}

With the rapid advancement of emerging technologies, autonomous driving is poised to transform future transportation systems by enhancing traffic safety~\cite{ye2019evaluating}, smoothing traffic flow~\cite{wang2022optimal}, and improving energy efficiency~\cite{sun2022energy}. As automated vehicles (AVs) featuring adaptive cruise control (ACC) become more common, they are anticipated to coexist with a large number of human-driven vehicles (HVs) in mixed-traffic environments in the foreseeable future~\cite{wang2022policy}. In these settings, the interactions between AVs and HVs are critical, significantly affecting traffic safety, stability, and energy consumption. However, many current AV control designs prioritize individual vehicle or platoon performance~\cite{wang2022optimal,aguilar2024simple}, potentially overlooking their wider impact on surrounding human drivers and the overall traffic dynamics.

The foundation for modeling these complex interactions lies in classical car-following (CF) theory. Seminal works, such as the cellular automation model~\cite{nagel1992cellular} and the nonlinear follow-the-leader model~\cite{cf_safety_comparison}, established the initial mathematical frameworks for modeling CF behavior. They have since been refined through empirical studies that reveal how human factors influence congestion and driving decisions~\cite{saifuzzaman2014incorporating}. Recent studies have adapted these classical approaches for vehicles equipped with ACC. For instance, new asymmetric CF models have been proposed to address the unique acceleration and deceleration characteristics of ACC systems~\cite{shang2022novel,shang2024two} and capture bidirectional velocity dependencies~\cite{gong2008asymmetric, pang2026large}. While robust, these analytical models often struggle to adapt to the variability and uncertainty inherent in real-world traffic.

To overcome these limitations, recent research has increasingly focused on physics-informed and learning-based models. By integrating physical CF principles into deep learning frameworks~\cite{mo2021physics} or employing multi-agent Long Short-Term Memory (LSTM) networks~\cite{li2024customizable}, these approaches can more accurately capture the temporal dependencies and latent cooperative patterns in driving behavior. Hybrid models that combine LSTMs with rule-based modules offer even greater generalizability~\cite{li2025racer}. These learning-based methods are further supported by empirical studies demonstrating behavioral adaptations when drivers transition between manual and ACC modes~\cite{varotto2020adaptations}, as well as by advanced calibration techniques that enable realistic trajectory fitting and personalization~\cite{desouza2020calibrating,nava2019personalized}. A summary of these key developments in ACC modeling and control is presented in Table~\ref{tab:key_lit}.

\begin{table*}[t!]
\caption{Summary of Key Literature on ACC Modeling and Control}
\centering
\small 
\setlength{\tabcolsep}{4pt} 
\begin{tabular}{@{}p{3.5cm}ccccp{5.5cm}p{3.5cm}@{}}
\toprule
\textbf{Study} & \textbf{Year} & \textbf{Model} & \textbf{Learning} & \textbf{Social} & \textbf{Key Innovation} & \textbf{Performance/Impact} \\
\midrule
\multicolumn{7}{l}{\textit{\textbf{Traditional Car-Following Models}}} \\
\midrule
Shang et al.~\cite{shang2022novel} & 2022 & \checkmark & & & Asymmetric OVRV (AOVRV) with deterministic ODE & 44.8\% spacing error reduction \\
De Souza \& Stern~\cite{desouza2020calibrating} & 2020 & \checkmark & & & Multi-objective Pareto analysis for OVRV/IDM/Gipps & Better spacing/speed tradeoffs \\
Shang et al.~\cite{shang2024two} & 2024 & \checkmark & & & Two-Condition Asymmetric CF (TCACF) & 36.98\% error reduction \\
Gong \& Liu~\cite{gong2008asymmetric} & 2008 & \checkmark & & & Asymmetric Full Velocity Difference (AFVD) & Wave dissipation captured \\
Zare et al.~\cite{zare2023modeling} & 2023 & \checkmark & & & EV-specific ACC car-following model calibrated with testbed data & Improved trajectory prediction accuracy \\
\midrule
\multicolumn{7}{l}{\textit{\textbf{Machine Learning Approaches}}} \\
\midrule
Mo et al.~\cite{mo2021physics} & 2021 & & \checkmark & & Physics-Informed Deep Learning (PIDL) & Robust to sparse noisy data \\
Nava et al.~\cite{nava2019personalized} & 2019 & & \checkmark & & Personalized ACC driving style characterization via learning & User-specific responses \\
Rhinehart et al.~\cite{rhinehart2019precog} & 2019 & & \checkmark & & PRECOG conditional intent modeling & Intent-aware AV planning \\
Dabiri \& Abbas~\cite{dabiri2018evaluation} & 2018 & & \checkmark & & GBRT model trained on NGSIM trajectory data & Outperforms classical GHR car-following model \\
\midrule
\multicolumn{7}{l}{\textit{\textbf{Control and Optimization}}} \\
\midrule
Wang et al.~\cite{wang2007safety_algorithms} & 2007 & \checkmark & & & ACC safety logic with risk envelopes & Safe decel/accel thresholds \\
Ozkan \& Ma~\cite{ozkan2022socially} & 2022 & \checkmark & & \checkmark & Socially compatible control design integrating SVO in car-following & Reduced disruption to human driver's original plan \\
\midrule
\multicolumn{7}{l}{\textit{\textbf{Social and Ethical Considerations}}} \\
\midrule
Shang et al.~\cite{shang2024interaction} & 2024 & \checkmark & & \checkmark & Interaction-aware MPC for AVs in mixed-autonomy traffic & Enhanced AV-HV interaction modeling \\
Wang~\cite{wang2024autonomous} & 2024 & \checkmark & & \checkmark & SVO-based optimal control & Energy efficiency in mixed traffic \\
Schwarting et al.~\cite{schwarting2019social} & 2019 & & \checkmark & \checkmark & IRL with SVO for dynamic games & Socially compliant behavior \\
Varga et al.~\cite{varga2025interaction} & 2025 & \checkmark & & \checkmark & IAMPDM with interaction awareness & Enhanced pedestrian safety \\
Dong et al.~\cite{dong2025towards} & 2025 & & & \checkmark & SCAV conceptual framework & Research gap identification \\
Liu et al.~\cite{liu2023social} & 2023 & & \checkmark & \checkmark & RL with driving priors and SCA & Socially responsive behaviors \\
\bottomrule
\end{tabular}
\label{tab:key_lit}
\end{table*}

While these advancements have improved model fidelity, a crucial challenge remains: optimizing AV behavior not solely for individual objectives but for the collective benefit of the entire traffic stream. This challenge has led to the emergence of \textit{socially compliant automated driving} as a vital subfield. The objective is to develop AVs that can interact harmoniously with human drivers and proactively regulate traffic flow. Recent frameworks have begun to explore this concept by incorporating Social Value Orientation (SVO), a theory from psychology that models the trade-off between self-interest and collective welfare, to balance individual and group benefits~\cite{ozkan2022socially}. This includes the development of control laws for fuel-efficient, harmonious driving~\cite{wang2024autonomous}, as well as reinforcement learning approaches that enable socially sensitive maneuvers~\cite{liu2023social}. These latest contributions signify a critical shift from purely technical AV planning toward behaviorally informed, human-centric decision-making.

Reinforcing the push toward socially compliant AVs, a range of recent studies contribute critical advancements across modeling, data, learning, and security. Makridis et al.~\cite{makridis2021openacc} introduced the OpenACC dataset, offering empirical car-following trajectories from commercial ACC systems, while Zhao et al.~\cite{zhao2020field} investigated human responses to AV behavior in longitudinal settings. To guide AVs toward human-aligned decisions, Murphy et al.~\cite{murphy2011measuring} developed behavioral scales for SVO, which inform how AV control logic can prioritize prosocial traffic integration. From a learning perspective, Mantouka and Vlahogianni~\cite{mantouka2022deep} used reinforcement learning to adapt aggressiveness levels of an AV based on human personality traits, while Ruan and Di~\cite{ruan2022learning} applied causal imitation learning to replicate realistic human trajectories. However, this increased behavioral predictability also introduces cyber vulnerabilities. Wang~\cite{wang2025analytical} demonstrated how attackers can exploit prosocial tendencies in AV controllers, and Wang et al.~\cite{wang2023optimal} proposed robust min-max control designs to mitigate these risks. These multidisciplinary contributions expand the scope of socially compliant driving, providing essential behavioral, technical, and security insights.

Despite this progress, there remains a lack of frameworks that integrate the predictive power of deep learning with the explicit social considerations of SVO for real-time AV control. This study addresses this critical gap by introducing \textit{Learn2Drive}, a novel neural network-based AV control framework that incorporates bidirectional CF dynamics into the control design process. By optimizing utility functions that reflect both AV-specific objectives and broader traffic efficiency goals through SVO, the proposed approach provides a socially compliant, adaptive, and efficient solution for harmonizing AV-HV coexistence in mixed traffic.

The main contributions of this study are threefold:
\begin{itemize}
    \item We propose a novel, socially compliant control framework, Learn2Drive, which integrates a neural network with physics-informed constraints and the SVO principle to effectively manage automated driving in mixed traffic environments.
    \item We formulate a unique SVO-based utility function that accounts for bidirectional CF dynamics and incorporate it into the proposed framework, enabling an AV to dynamically balance its own energy efficiency with the driving stability of the following HV.
    \item We analyze the bidirectional interactions between an AV and its following HV, demonstrating how socially attuned AVs can act as mobile traffic regulators to reduce congestion and improve overall traffic efficiency.
\end{itemize}

The remainder of this article is organized as follows. Section~\ref{sec:model} details the proposed Learn2Drive framework, including the underlying CF models. Section~\ref{sec:data} describes the dataset used for training and validation, while Section~\ref{sec:experiments} outlines the numerical experiments conducted. In Section~\ref{sec:analysis}, we present and discuss the results and their implications. Finally, Section~\ref{sec:con} concludes the article with a summary of our findings, a discussion of limitations, and potential avenues for future research.

\section{Methodology}\label{sec:model}

\begin{figure}[t!]
    \centering
    \includegraphics[width=0.5\textwidth]{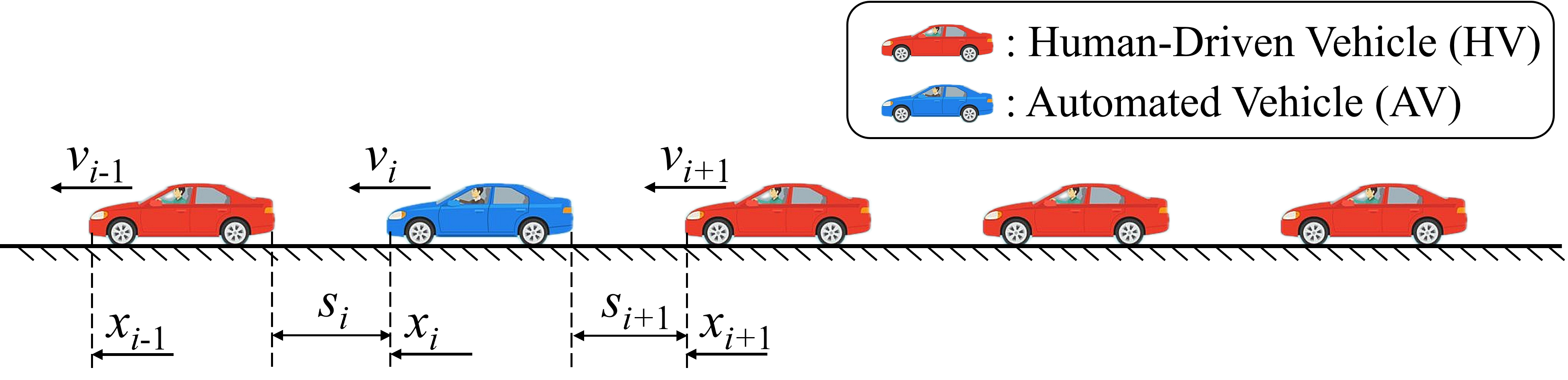}
    \caption{Illustration of traffic flow with AVs and HVs, highlighting the role of a single AV as a key coordinator to guide traffic movement, reduce congestion, and enhance system efficiency.}
    \label{fig:traffic_flow_AV_HV}
\end{figure}

This study presents Learn2Drive, a forward-looking approach for socially compliant AV control, designed to enhance traffic flow efficiency and promote sustainable outcomes in mixed traffic environments. The central aim of Learn2Drive is to achieve a balanced integration of individual vehicle performance with the collective benefits of the broader traffic system, reimagining transportation as a cooperative network. The proposed framework is built on the core principle of using a single AV as a key coordinator to guide traffic movement, alleviate congestion, and improve overall system reliability, as illustrated in Fig.~\ref{fig:traffic_flow_AV_HV}. This AV operates without requiring communication with other vehicles, relying solely on local observations such as spacing and relative speeds captured by onboard sensors to make socially compliant decisions. By leveraging these self-reliant observations, the AV achieves coordination without vehicle-to-vehicle communication or external data dependencies, ensuring robust performance in diverse traffic scenarios. The methodology unfolds through three interconnected stages: (1) a generalized traffic dynamics model, (2) utility-based optimization incorporating social preferences, and (3) adaptive learning with a feedback mechanism, progressing from broad conceptual insights to a structured, implementable approach for advancing intelligent transportation systems.

\subsection{Generalized Traffic Dynamics Model}

Learn2Drive starts with a generalized traffic dynamics model, a foundational concept that explores the interactive movement of vehicles within a traffic system. This model begins with the broad understanding that vehicle interactions shape the flow and structure of traffic, proposing the use of an AV as a central guide to influence the entire traffic. The design philosophy of Learn2Drive hinges on the innovative concept of employing a single AV to control the entire traffic flow, a principle rooted in the belief that a strategically positioned intelligent vehicle can serve as a catalyst for systemic improvement~\cite{cui2017stabilizing, wu2018stabilizing}. This approach is based on the idea that the AV, through its advanced sensing and decision-making capabilities, can act as a stabilizing force, proactively adjusting its behavior to mitigate congestion, smooth traffic patterns, and enhance overall system efficiency. This decentralized control paradigm transforms the traffic system into a self-regulating entity, where the actions of one AV ripple through the network, fostering a balanced and efficient flow. This principle is distilled into a set of governing equations that outline the dynamics:
\begin{equation}
\dot{x}_i(t) = v_i(t)
\end{equation}
where \( \dot{x}_i(t) \) represents the rate of change of vehicle \( i \)'s position, and \( v_i(t) \) is its speed, establishing a foundation for tracking motion. The acceleration of vehicle \( i \) is given by~\cite{wilson2011car}:
\begin{equation}
\dot{v}_i(t) = f(s_i(t), \Delta v_i(t), v_i(t))
\end{equation}
where \( s_i(t) \) is the spacing between vehicle \( i \) and its predecessor, \( \Delta v_i(t) \) is the relative speed, and \( v_i(t) \) is the vehicle's own speed, reflecting the interactive nature of traffic dynamics. Specifically, \( \Delta v_i(t) \) is defined as:
\begin{equation}
\Delta v_i(t) = \dot{s}_i(t) = v_{i-1}(t) - v_i(t)
\end{equation}

These equations provide traffic dynamics that support a variety of traffic patterns, reinforcing Learn2Drive's goal of building a self-regulating traffic system. To account for the mixed traffic environment comprising both AVs and HVs, we extend the model to differentiate their dynamics. Let \( \mathcal{V} = \mathcal{P} \cup \mathcal{Q} \) denote the set of all vehicles, where \( \mathcal{P} \) and \( \mathcal{Q} \) represent the ordered sets of HVs and AVs, respectively. Human drivers exhibit distinct behavioral traits compared to AVs, such as variability in reaction times and decision-making~\cite{saifuzzaman2014incorporating}. To capture these differences, the acceleration dynamics are further specified as:
\begin{equation}
\dot{v}_i(t) = 
\begin{cases} 
f_{\text{HV}}(s_i(t), \Delta v_i(t), v_i(t)), & i \in \mathcal{P} \\
f_{\text{AV}} = \text{NN}(s_i(t), \Delta v_i(t), v_i(t)), & i \in \mathcal{Q}
\end{cases}
\end{equation}
where \( f_{\text{HV}}(s_i(t), \Delta v_i(t), v_i(t)) \) is a traditional physical car-following model that captures empirical driving behaviors of HVs based on spacing \( s_i(t) \), relative speed \( \Delta v_i(t) \), and current speed \( v_i(t) \), and \( \text{NN}(s_i(t), \Delta v_i(t), v_i(t)) \) is an AI model, specifically a neural network, that predicts the AV’s acceleration using the same input parameters, designed to incorporate socially compliant objectives within the Learn2Drive framework. These functions account for the distinct driving characteristics of HVs and AVs, enabling the model to reflect the heterogeneous nature of mixed traffic.

The control of the entire traffic flow is achieved through the strategic modulation of the AV's acceleration, \( f_{\text{AV}} \), which is directly optimized using a neural network-based approach to balance individual performance and collective traffic objectives. Learn2Drive computes \( f_{\text{AV}} \) based on real-time traffic data, such as spacing \( s_i \), relative speed \( \Delta v_i \), and the AV's own speed \( v_i \), which are readily available for ACC systems~\cite{wang2023general}. By dynamically adjusting its acceleration, the AV acts as a traffic regulator, influencing the behavior of following HVs to mitigate congestion, stabilize traffic flow, and enhance system-wide efficiency. This approach leverages the AV's advanced sensing and decision-making capabilities to propagate stabilizing effects throughout the traffic flow, aligning with Learn2Drive's vision of a self-regulating, socially compliant transportation system.

\subsection{Utility-Based Optimization with Social Preferences}

Emerging from the traffic dynamics model and its design philosophy of using a single AV to regulate the entire traffic flow, Learn2Drive refines this concept by focusing on a novel control strategy that addresses the limitations of traditional approaches and incorporates social preferences to enhance traffic system performance. This subsection outlines the challenges of conventional AV control, introduces SVO as a solution, and details how SVO is applied to achieve socially compliant traffic flow control.

Traditional AV control strategies, such as those used in ACC systems, often prioritize individual objectives, such as minimizing energy consumption or optimizing travel time. While effective in isolated scenarios, these approaches can produce behaviors that are unpredictable or disruptive in mixed-traffic environments. For instance, an AV focused solely on its own efficiency may accelerate or decelerate abruptly, inducing stop-and-go waves that frustrate human drivers, increase fuel consumption, and destabilize traffic flow. Such self-centered control overlooks the cooperative dynamics essential in mixed traffic, where human drivers rely on predictable and intuitive interactions to ensure safety and efficiency. This misalignment can exacerbate congestion, diminish traffic stability, and even compromise safety, underscoring the need for a control framework that accounts for the broader impact on surrounding HVs and the overall traffic system.

To address these challenges, Learn2Drive incorporates SVO, a psychological concept that quantifies the balance between self-interest and collective welfare~\cite{murphy2011measuring}, as a guiding principle for AV control design. SVO enables an AV to weigh its own performance objectives against the needs of surrounding HVs, fostering cooperative and legible behavior that aligns with human social expectations~\cite{wang2024autonomous}. By modeling preferences along a spectrum from egoistic (prioritizing self-interest) to prosocial (balancing individual and collective benefits) to altruistic (prioritizing others' welfare), SVO allows the AV to adapt its actions to promote a harmonious traffic environment. This approach is critical in mixed traffic, where cooperation between AVs and HVs is essential for enhancing safety, reducing congestion, and improving system efficiency. By integrating SVO, Learn2Drive ensures that the AV acts as a traffic coordinator, proactively adjusting its behavior to benefit the entire traffic, aligning with the framework's vision of a self-regulating, sustainable transportation system.

SVO is applied to traffic flow control by defining a utility function that optimizes the AV's acceleration to balance individual and collective objectives, using real-time traffic data to inform decision-making. The utility function weighs the AV's performance metrics, such as energy efficiency, against collective goals, such as stabilizing HV speeds or maintaining safe spacing. A social preference parameter, \( \phi \), determines the degree of emphasis on individual versus collective benefits, allowing the AV to dynamically adjust its behavior based on dynamic traffic conditions~\cite{wang2024autonomous}.
The utility function is defined as:
\begin{equation}
U = \cos(\phi) \cdot U_{\text{self}} + \sin(\phi) \cdot U_{\text{collective}}
\label{eq:utility}   
\end{equation}
where $\phi \in [0, \pi/2]$ parameterizes the AV's social preference, following the standard geometric formulation in SVO theory. In this representation, $\phi = 0$ corresponds to purely egoistic behavior (maximizing self-interest), $\phi = \pi/2$ to purely altruistic behavior (maximizing collective welfare), and intermediate values (e.g., $\phi = \pi/4$) represent prosocial trade-offs. The trigonometric weights $\cos(\phi)$ and $\sin(\phi)$ are naturally bounded and continuous and satisfy a \emph{unit-circle normalization} ($\cos^2\phi+\sin^2\phi=1$) across the full preference range, providing a psychologically grounded and geometrically interpretable angle-based parameterization. While a linear combination with $\alpha$ and $(1-\alpha)$ is also possible, it does not endow $\phi$ with the same directional interpretation in the SVO plane; if coefficients summing to one are desired under an angle parameterization, a squared reparameterization (e.g., $\cos^2\phi$ and $\sin^2\phi$) can be used. We adopt the standard angle-vector form here to preserve the canonical SVO interpretation and continuity across $\phi$.

The control strategy is implemented through a neural network-based approach that processes inputs like inter-vehicle spacing, relative speeds, and the AV's own speed to compute an optimal acceleration that maximizes $U$. For example, a greater \( \phi \) in dense traffic prompts smoother acceleration patterns to reduce stop-and-go waves, benefiting HVs by stabilizing their speeds and improving overall traffic flow. Conversely, a smaller \( \phi \) in free-flow conditions allows the AV to prioritize its own efficiency, such as minimizing energy consumption, while still maintaining predictable interactions to ensure safety for HVs. By modulating \( \phi \) based on contextual factors like traffic density or congestion levels, the AV acts as a traffic regulator, influencing HV behavior to mitigate congestion, stabilize flow, and enhance system efficiency. Learn2Drive's design underscores a commitment to embedding social awareness into vehicular systems, creating a transportation network where individual and collective goals are seamlessly aligned, fostering a sustainable traffic future through socially compliant AV control.

\subsection{Adaptive Learning with Feedback Mechanism}

Building on the SVO-guided optimization, Learn2Drive advances to an adaptive learning mechanism with feedback improvement, addressing the unique challenges posed by AV behavior in mixed traffic environments. This component starts with the recognition that AVs, with their advanced sensing and decision-making capabilities, require a principle-based, data-supported approach to effectively manage complex traffic dynamics. This leads to the adoption of AI models to predict and control AV acceleration, replacing conventional car-following models that struggle to capture the nuanced interactions in diverse traffic conditions. The AI approach offers a principle-driven alternative, leveraging data to model acceleration patterns more effectively, ensuring accurate representation of AV behavior across varying scenarios. The learning process is guided by a loss function comprising \( \mathcal{L}_{\text{prediction}} \) and \( \mathcal{L}_{\text{cost}} \), reflecting the system's ability to adapt to traffic variations. The learning process is guided by:
\begin{equation}
\label{eq:loss}
\mathcal{L} = \mathcal{L}_{\text{prediction}} + \mathcal{L}_{\text{cost}}
\end{equation}
where \( \mathcal{L} \) combines the prediction and utility components to guide learning. Specifically, 

\begin{equation}
\mathcal{L}_{\text{prediction}} = g(\mathbf{y}_{\text{model}}, \mathbf{y}_{\text{obs}})
\end{equation}
where \( \mathcal{L}_{\text{prediction}} \) aligns modeled states \( \mathbf{y}_{\text{model}} \) with observed states \( \mathbf{y}_{\text{obs}} \) through a function \( g \). Further,
\begin{equation}
\mathcal{L}_{\text{cost}} = h(U)
\end{equation}
where \( \mathcal{L}_{\text{cost}} \) optimizes traffic objectives through a function \(h \) linked to the utility \( U \).

The adaptation is influenced by contextual factors such as traffic conditions, enabling Learn2Drive to respond to changing traffic conditions by dynamically adjusting its control strategies to maintain traffic stability and efficiency. This mechanism enhances the system's ability to integrate real-time data with predictive modeling, ensuring a balanced approach to traffic management. It provides a principled foundation for intelligent, socially responsive transportation, supported by conceptual insights~\cite{mo2021physics}. Learn2Drive's design envisions a system that continuously learns from its environment to achieve collective resilience and adaptability over isolated performance. This integrated approach within Learn2Drive establishes a comprehensive framework for socially compliant AV control, addressing transportation challenges through the synergy of traffic dynamics, SVO-based optimization, and adaptive learning.

\section{Data Description}\label{sec:data}

Building upon the methodology outlined in the previous section, which introduced physical car-following models, SVO-guided utility optimization, and the Learn2Drive framework integrating a general AI approach to design the AV model, this section details the dataset used to evaluate the proposed approach. The specific architecture of the AI model employed in the experiments, including its selection and implementation, will be thoroughly described in Section~\ref{sec:experiments}. Traditional ACC car-following datasets, such as those from two-vehicle testbeds, typically involve only two vehicles and fail to reflect the complex dynamics of multi-vehicle traffic flow. The HighD dataset, which records many vehicles on German highways using drone-based tracking, offers trajectories limited to approximately 420 meters, making it insufficient for studying long-term traffic dynamics in closed-loop scenarios~\cite{krajewski2018highd}. Similarly, the NGSIM dataset, comprising vehicle trajectories from US highways like US-101 captured over short segments, provides brief individual vehicle paths, limiting its utility for analyzing sustained congestion effects~\cite{ngsim2016}. In contrast, the Arizona Ring Experiments Dataset (ARED)~\cite{wu2019tracking} was selected for its ability to provide extended trajectories in a controlled ring environment, effectively capturing multi-vehicle interactions in phantom traffic jams. ARED is a valuable resource compiled through experimental studies focused on traffic dynamics. In July 2016, researchers conducted eight experiments in Tucson, Arizona, to develop a method for extracting high-quality trajectory and fuel consumption data of vehicles in phantom traffic jams and to investigate how a single vehicle, simulating a low penetration rate on a freeway, can influence the onset and progression of these jams by driving differently from the surrounding traffic stream~\cite{wu2019tracking}. For this study, empirical trajectories from the dataset are utilized to establish a realistic leading vehicle profile and calibrate the human-driven follower models. This calibrated foundation enables the robust analysis of socially compliant AV control strategies within the Learn2Drive framework across various mixed-traffic configurations, ranging from a baseline 5-vehicle platoon to extended 8- and 10-vehicle formations.

The selection of ARED is motivated by its ability to capture essential aspects of traffic behavior, offering a broad foundation for evaluating the impact of individual vehicle actions on the overall system. This dataset aligns with Learn2Drive's goals of modeling and optimizing traffic interactions, providing a suitable basis for exploring innovative control strategies in mixed traffic environments. The experimental design of ARED, focused on real-world traffic phenomena, supports the framework's objectives by offering insights into the dynamic interactions influenced by a single AV's behavior.

\begin{figure}[t!]
    \centering
    \subfloat[Speed profiles of the first five vehicles]{\includegraphics[width=0.4\textwidth]{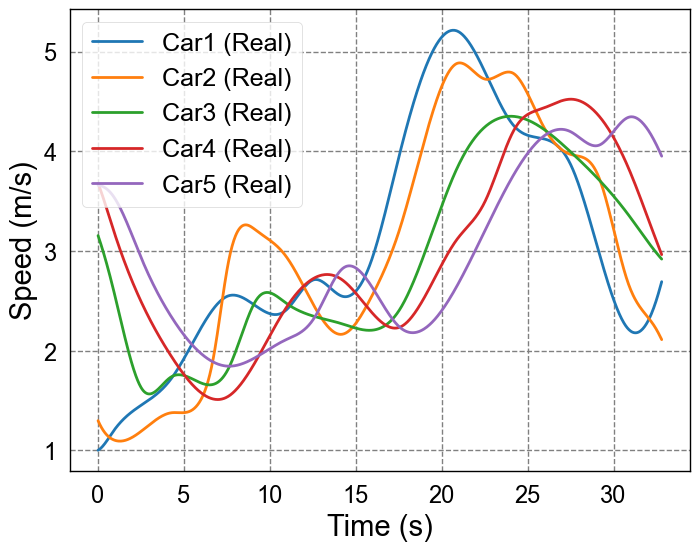}\label{fig:speed_trajectory}} 
    \vfill
    \subfloat[Spacing profiles of the first five vehicles]{\includegraphics[width=0.4\textwidth]{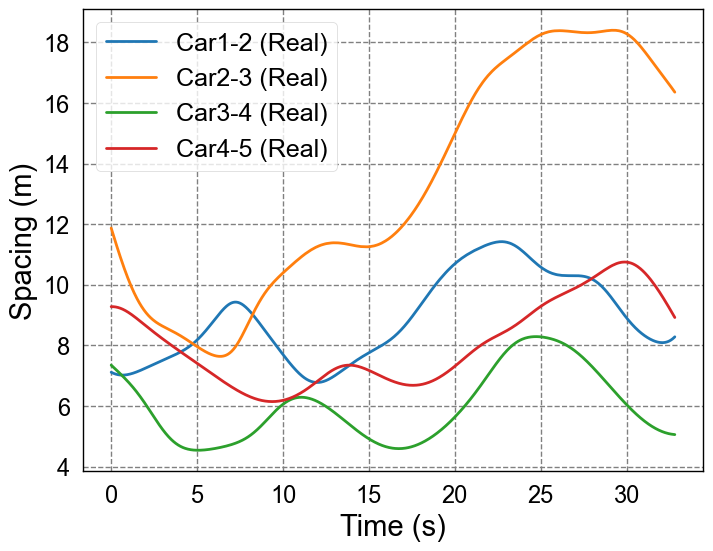}\label{fig:spacing_trajectory}}
    \caption{Partial profiles of speed and inter-vehicle spacing for the first five vehicles in the ARED dataset, illustrating the platoon's car-following dynamics.}
    \label{fig:trajectories}
\end{figure}

The trajectory data were extracted from 360-degree video footage using a tracking-by-matching method, which involves background subtraction and object identification to reconstruct position, velocity, and spacing over time. A fourth-order weighted B-spline smoother was applied to the raw data to reduce noise caused by varying lighting conditions or reflective surfaces, enhancing the accuracy of velocity and position estimates. The dataset includes time-series measurements of vehicle speed and inter-vehicle spacing, with dynamic interactions visualized in Fig.~\ref{fig:trajectories}, which highlights the platoon's behavior under varying traffic conditions.

These trajectory data provide critical insights into car-following dynamics, enabling the evaluation of the Learn2Drive framework's performance in optimizing AV control under different SVO settings. Validation of the dataset's accuracy was conducted by comparing smoothed camera velocity estimates with OBD-II data, achieving consistency as reported in~\cite{wu2019tracking}. This subset supports the analysis of speed and spacing root mean square error (RMSE), AV energy consumption, and HV speed deviations, which will be quantitatively assessed in the following sections.

\section{Experiments}\label{sec:experiments}
This section describes the closed-loop simulation used to evaluate Learn2Drive. The baseline setting is a 5-vehicle platoon based on the ARED ring-road dataset. The lead vehicle $v_1$ follows the measured trajectory, the controlled AV $v_2$ is driven by the learned LSTM policy, and the follower vehicles $v_3$--$v_5$ are modeled by calibrated IDM controllers. Vehicle states are updated by forward Euler integration with $\Delta t = 1/30$ s. Additional 8-vehicle and 10-vehicle experiments are introduced later to assess whether the observed SVO-driven effects persist beyond the baseline platoon.

\subsection{Experimental Framework}

The numerical experiments are methodically engineered to simulate a platoon of vehicular entities, providing a concrete realization of the generalized traffic dynamics model by replicating the interdependent motion within a traffic flow. The entity \(v_1\), designated as the leading vehicle, is propelled by empirical trajectory data extracted from the 2016 Tucson experiments~\cite{wu2019tracking}, establishing a dependable reference that reflects real-world driving behaviors and anchors the simulation in observed patterns. The entity \(v_2\), configured as the controlled AV, adapts its actions through a learned control policy informed by the LSTM network and SVO-based optimization, serving as the central focus for evaluating socially compliant strategies and fulfilling the framework's objective of influencing traffic flow modalities. The entities \(v_3\), \(v_4\), and \(v_5\), acting as follower vehicles, are modeled using a conventional physics-based method to simulate the responses of HVs, enabling the examination of traffic flow reactions, such as consistent movement or disrupted patterns, stemming from the AV's influence. This configuration operationalizes the concept of interconnected motion through a discrete-time evolution of states, where speed \(v_i\) and spacing \(s_i\) for each vehicular entity \(i\) are updated as:
\begin{equation}
v_i(t + \Delta t) = v_i(t) + a_i(t) \Delta t
\end{equation}
where \(v_i(t)\) represents the speed of vehicle \(i\) at time \(t\), \(a_i(t)\) is the acceleration, and \(\Delta t\) is a small time step; this equation captures the continuous change in speed based on acceleration, providing an update mechanism for the simulation. In addition, 
\begin{equation}
s_i(t + \Delta t) = s_i(t) + [v_{i-1}(t) - v_i(t)] \Delta t
\end{equation}
where \(s_i(t)\) denotes the spacing between vehicle \(i\) and its predecessor, \(v_{i-1}(t)\) is the speed of the preceding vehicle, and \(v_i(t)\) is the speed of vehicle \(i\); this equation reflects the relative motion affecting inter-vehicle spacing, ensuring accuracy by considering the immediate predecessor's speed, consistent with the simulation's dependency on lead vehicle dynamics. The computational platform enables these simulations across a diverse set of traffic conditions, employing synthesized data informed by the Tucson dataset to reflect temporal changes and a broad spectrum of driving scenarios. The experimental setup is guided by the principle of creating a controlled environment that mirrors real-world traffic interactions, allowing for the systematic evaluation of how the controlled entity's behavior influences the collective dynamics of the platoon, with the ultimate goal of validating the AV's role as a modulator of traffic flow patterns.

\subsection{Traffic Flow Simulation for a Controlled AV}\label{subsec:traffic_flow_av}

The algorithmic formulation for the controlled AV, operationalizes the adaptive learning mechanism, offering a concrete manifestation of the utility-based optimization and feedback-driven refinement articulated in Section~\ref{sec:model}. This approach utilizes an LSTM network to predict the AV control input \( u(t) \), i.e., its acceleration \( a_{\text{AV}}(t) \), optimized across various social preference settings, \( \phi \in \{0, \pi/4, \pi/2\} \). The control input is computed as:
\begin{equation}
u(t) = a_{\text{AV}}(t) = f_{\text{LSTM}}(s_{1,2}(t), \Delta v_{1,2}(t), v_2(t), \phi)  \label{eq_control}
\end{equation}
where \( f_{\text{LSTM}} \) represents the LSTM network function, \( s_{1,2} \) is the spacing between the AV and its lead vehicle, \( \Delta v_{1,2} = v_1 - v_2 \) is the relative speed, \( v_2 \) is the AV's speed, and \( \phi \) is the social preference parameter guiding the balance between individual and collective objectives. 
It is noted that the AVs are modeled as non-connected vehicles without vehicle-to-vehicle (V2V) or vehicle-to-infrastructure (V2I) communication capabilities. Therefore, they cannot access information beyond the immediate predecessor, such as multi-vehicle states ahead or behind, traffic signals, road geometry, or network-level data. The optimization in equation~\eqref{eq_control} uses only the same local variables (spacing $s_i$, relative speed $\Delta v_i$, and own speed $v_i$) as in conventional car-following models to reflect realistic constraints of standard ACC sensors (e.g., radar or LiDAR typically providing only immediate predecessor information). This design serves as a proof-of-concept to demonstrate the effectiveness of SVO-guided control under current single-vehicle perception capabilities.

The acceleration of the controlled AV ($v_2$) is predicted by a LSTM network, which is configured with the following general architecture and hyperparameters:

\begin{itemize}
\item Architecture: single LSTM layer with 64 hidden units, input dimension 10 (detailed in Section IV-A for feature engineering), followed by dropout and ReLU-activated feed-forward layers, with output scaled by $\tanh$ and $\phi$-dependent clamping to $[-0.6, 1.0]$ m/s$^2$;
\item Optimizer: AdamW with adaptive learning rate scheduling;
\item Training setup: chronological train/test split, sequence length 15, batch size 64, gradient clipping, and early stopping based on validation loss;
\item Convergence criteria: stabilization of validation loss and satisfaction of SVO trend constraints ($U_{\mathrm{self}}$ non-decreasing and $U_{\mathrm{collective}}$ non-increasing across $\phi$ values).
\end{itemize}

These specifications ensure stable training and reproducible results. A summary of the main hyperparameters is provided in Table~\ref{tab:lstm_hyperparams}.

\begin{table}[t!]
\caption{Hyperparameters of the LSTM-based AV Controller}
\centering
\small
\begin{tabularx}{\columnwidth}{>{\centering\arraybackslash}p{3.8cm}>{\centering\arraybackslash}X}
\toprule
\textbf{Parameter} & \textbf{Value} \\
\midrule
LSTM layers                  & 1 \\
Hidden units                 & 64 \\
Dropout rate                 & 0.1 \\
Input size                   & 10 \\
Sequence length              & 15 \\
Batch size                   & 64 \\
Optimizer                    & AdamW \\
Learning rate scheduling     & Adaptive \\
Gradient clipping            & Applied \\
Early stopping               & Based on validation loss \\
Train/test split             & Chronological \\
Convergence criteria         & Loss stabilization + SVO trend satisfaction \\
\bottomrule
\end{tabularx}
\label{tab:lstm_hyperparams}
\end{table}

The LSTM network is chosen for its effectiveness in modeling sequential dependencies within traffic data, capturing nonlinear relationships, and adapting to varying traffic conditions, making it well-suited for representing the temporal intricacies of car-following behavior. The prediction of \( u(t) \) focuses on longitudinal acceleration, enabling dynamic adjustments to traffic conditions through learning from time-series data over a specified sequence length, allowing the AV to modulate its acceleration to balance individual performance and collective traffic objectives as dictated by the SVO framework. The learning process is directed by a loss function given by:
\begin{equation}
\mathcal{L} = \alpha \mathcal{L}_{\text{prediction}} + \beta \mathcal{L}_{\text{cost}} + \gamma \mathcal{L}_{\text{constraint}}
\end{equation}
where \(\mathcal{L}_{\text{prediction}}\) ensures prediction accuracy, \(\mathcal{L}_{\text{cost}}\) drives utility optimization, and \(\mathcal{L}_{\text{constraint}}\) enforces operational constraints, with weighting factors \(\alpha\), \(\beta\), and \(\gamma\) refining the optimization process. This equation represents the total loss \(\mathcal{L}\), extending the foundational principle outlined in Section~\ref{sec:model} (see Equation~\eqref{eq:loss}). Equation~\eqref{eq:loss} presents the general design of the loss function during the methodology stage, while this equation provides its specific implementation in the experimental section by introducing an additional constraint term to ensure reasonable car-following behavior and adjusting the weights of different loss terms to address their magnitude differences. Specifically, \(\alpha\) emphasizes the accuracy of acceleration predictions, \(\beta\) prioritizes alignment with utility-based social preferences, and \(\gamma\) ensures adherence to operational constraints, allowing for a tailored balance across these objectives within the traffic flow simulation, where \(\mathcal{L}_{\text{constraint}}\) is composed of traffic smoothness (\(\mathcal{L}_{\text{smoothness}}\)) and trend consistency (\(\mathcal{L}_{\text{trend}}\)) components.

To prevent any single component from dominating the optimization due to magnitude differences, we adopt two key balancing strategies in the loss formulation.
First, the raw utility terms $U_{\text{self}}$ (energy-related, typically small magnitude) and $U_{\text{collective}}$ (speed deviation-related, often larger magnitude) in $\mathcal{L}_{\text{cost}}$ (Eqs.~\eqref{eq:norm_1}-\eqref{eq:norm_3}) are normalized using dataset-specific scaling factors before applying the SVO weights. The scaling factors $s_{\text{self}}$ and $s_{\text{collective}}$ are computed from the training dataset as the mean absolute values of each utility term over all samples:
\begin{equation}
s_{\text{self}} = \frac{1}{N} \sum_{i=1}^{N} |U_{\text{self}}^{(i)}|, \quad s_{\text{collective}} = \frac{1}{N} \sum_{i=1}^{N} |U_{\text{collective}}^{(i)}|,
\label{eq:norm_1}
\end{equation}
where $N$ is the number of training samples. The normalized utilities are then obtained as:
\begin{equation}
\tilde{U}_{\text{self}} = \frac{U_{\text{self}}}{s_{\text{self}}}, \quad \tilde{U}_{\text{collective}} = \frac{U_{\text{collective}}}{s_{\text{collective}}}.
\label{eq:norm_2}
\end{equation}
The SVO cost term becomes:
\begin{equation}
\mathcal{L}_{\text{svo}} = \cos(\phi) \cdot \tilde{U}_{\text{self}} + \sin(\phi) \cdot \tilde{U}_{\text{collective}}
\label{eq:norm_3}
\end{equation}
This normalization ensures that both terms contribute on a comparable scale regardless of their raw physical units or typical value ranges.
Second, to further maintain balance across all loss components ($\mathcal{L}_{\text{follow}}$, $\mathcal{L}_{\text{svo}}$, $\mathcal{L}_{\text{smoothness}}$, $\mathcal{L}_{\text{trend}}$), we employ an adaptive weighting strategy during training. In each epoch, the weights $\alpha$, $\beta$, $\gamma$ (and internal sub-weights) are dynamically adjusted based on the relative magnitudes of the individual loss terms observed in the previous epoch. Specifically, each weight is scaled inversely proportional to the average loss value of that component in the prior epoch (with a small minimum threshold to avoid zero weights), so that larger loss terms receive relatively smaller weights and vice versa. This adaptive mechanism is updated at the beginning of each epoch and helps prevent early dominance by any one term while allowing natural convergence behavior as training progresses.
These normalization and adaptive weighting strategies collectively ensure that the multi-objective optimization remains balanced across different social preference values $\phi$, leading to more stable training dynamics and better alignment with the intended SVO trade-offs.

Specifically, $\mathcal{L}_{\text{prediction}}$ is given by:
\begin{equation}
\mathcal{L}_{\text{prediction}} = \int_{0}^{T} \left( a_{\text{pred}}(t) - a_{\text{true}}(t) \right)^2 \, dt
\end{equation}
where \(T\) is the time horizon, \(a_{\text{pred}}(t)\) is the predicted acceleration, and \(a_{\text{true}}(t)\) is the true acceleration; this term ensures the predicted acceleration aligns with real data by penalizing the squared difference over time, promoting accurate tracking of vehicle dynamics. {$a_{\text{true}}$ denotes the ground-truth acceleration of the AV from the ARED dataset (used as the supervised training target); $a_{\text{pred}}$ denotes the acceleration predicted by the LSTM model; and $a_{\text{AV}}$ denotes the acceleration applied to the AV in the closed-loop simulation, which is the output of the LSTM model (i.e., $a_{\text{AV}} = a_{\text{pred}}$). The prediction loss minimizes the difference between $a_{\text{pred}}$ and $a_{\text{true}}$ to ensure the model learns realistic driving behavior.
$\mathcal{L}_{\text{cost}}$ is defined as:

\begin{equation}
\mathcal{L}_{\text{cost}} = \int_{0}^{T} \left[ \cos(\phi) \cdot \tilde{U}_{\text{self}}(t) + \sin(\phi) \cdot \tilde{U}_{\text{collective}}(t) \right] \, dt
\end{equation}
where \( \phi \) is the social preference parameter, and \( U_{\text{self}} \) and \( U_{\text{collective}} \) are utility terms; this term balances individual and collective objectives across different values of \( \phi \), with the contribution to the total loss reflecting the minimization of utility, encouraging efficient energy use and stable traffic flow.

Following our recent study~\cite{wang2024autonomous}, the utility terms are specifically given by:
\begin{equation}
U_{\text{self}}(t) = \int_{0}^{T} \frac{1}{2} a_{\text{AV}}^2(t) \, dt
\end{equation}
which represents the AV's energy consumption, serving as a measure of individual performance to be minimized. Moreover, 

\begin{equation}
U_{\text{collective}}(t) = \int_{0}^{T} \frac{1}{2} (v_{\text{i}}(t) - v_0)^2 \, dt
\end{equation}
where \( v_{\text{i}} \) is the speed of the follower vehicle, and \( v_0 \) is the desired speed from the CF model parameters; this quantifies the speed deviation from the target, aiming to improve traffic flow efficiency by reducing deviations over time.

The last component of $\mathcal{L}$ is given by:
\begin{equation}
\mathcal{L}_{\text{constraint}} = \mathcal{L}_{\text{smoothness}} + \mathcal{L}_{\text{trend}}
\end{equation}
where \(\mathcal{L}_{\text{constraint}}\) combines traffic smoothness and trend consistency; this term integrates smoothness and trend constraints to ensure stable and consistent driving behavior. Specifically.
\begin{equation}
\mathcal{L}_{\text{smoothness}} = \int_{0}^{T} \left( \frac{d a_{\text{AV}}(t)}{dt} \right)^2 \, dt
\end{equation}
where \(\frac{d a_{\text{AV}}(t)}{dt}\) is the jerk (first derivative of acceleration). This term enforces smooth acceleration profiles by penalizing rapid changes in acceleration, enhancing driving comfort and stability. The term $\mathcal{L}_{\text{trend}}$ is given by: 
\begin{align}
\mathcal{L}_{\text{trend}} = \int_{0}^{T} \Big[ &\left( U_{\text{self}}(t, \phi_1) - U_{\text{self}}(t, \phi_2) \right)^2 \notag \\
                            &\hskip-5pt + \left( U_{\text{collective}}(t, \phi_2) - U_{\text{collective}}(t, \phi_1) \right)^2 \Big] \, dt
\label{eq:trend_loss}
\end{align}
where \( \phi_1 \) and \( \phi_2 \) are different social preferences (\( \phi_1 < \phi_2 \)). This term enforces consistent utility trends across variations in \( \phi \) by penalizing deviations from expected patterns in individual and collective utilities. Specifically, it encourages \( U_{\text{self}} \), representing the AV's energy consumption, to increase with \( \phi \) (as the AV prioritizes collective benefits over individual efficiency) and \( U_{\text{collective}} \), reflecting HV speed stability, to decrease with \( \phi \) (as the AV enhances traffic flow). The loss is dynamically weighted during training to strengthen enforcement as learning progresses, ensuring that the AV's behavior adapts predictably across egoistic (\( \phi \approx 0 \)), prosocial (\( \phi \approx \pi/4 \)), and altruistic (\( \phi \approx \pi/2 \)) settings, promoting stable and socially compliant control. Equation~\eqref{eq:trend_loss} is used to encode the intended monotonic dependence of the utilities on the social-preference angle $\phi$, namely that $U_{\text{self}}$ is nondecreasing and $U_{\text{collective}}$ is nonincreasing as $\phi$ increases. In this study, the social-preference grid is discrete and fixed to $\{0,\ \pi/4,\ \pi/2\}$. During training, equation~(\ref{eq:trend_loss}) is evaluated pairwise across ordered angles and summed over $(\phi_1,\phi_2)$ with $\phi_1<\phi_2$, primarily using adjacent pairs $(0,\pi/4)$ and $(\pi/4,\pi/2)$ (optionally augmented by the endpoints pair $(0,\pi/2)$) to balance coverage and computational cost. In practice, equation~(\ref{eq:trend_loss}) is applied to residuals that indicate potential violations of the intended order, so the contribution is negligible when monotonicity is already satisfied with a nontrivial gap; this avoids unintended flattening. All utilities are first normalized, and the trend term is combined with the prediction, SVO, and smoothness components via adaptive weights to prevent early dominance and to stabilize multi-objective training.

The framework does not aim to directly align AV and HV accelerations for imitation or to characterize social awareness through behavioral matching. Instead, it optimizes the AV acceleration via the LSTM model using an explicit SVO-weighted utility that balances self-interest ($U_{\text{self}}$) and collective welfare ($U_{\text{collective}}$). The parameter $\phi$ is designed as an explicit, interpretable control knob for social preference, rather than a latent value learned from data. Directly learning $\phi$ (or the equivalent IDM parameter $\theta$) from human driver datasets would be challenging, as such datasets lack quantitative labels for egoistic versus altruistic behavior during natural driving. This explicit design enables interpretable tuning of the self-collective trade-off and proactive influence on upstream traffic, which is the fundamental goal of the proposed socially compliant control.

The LSTM network learns diverse AV driving behaviors (e.g., egoistic at \( \phi = 0 \) and altruistic at \( \phi = \pi/2 \)) by minimizing \( \mathcal{L} \) over training iterations, where variations in \( \phi \) guide the network to adapt its predictions, realizing control patterns such as stable following or cooperative spacing. A summary of this algorithm is provided in \textbf{Algorithm 1}, employing symbolic notation to reflect the framework's design.

\begin{table}[t!]
    \centering
    \caption*{\textbf{Algorithm 1}: Learn2Drive for Socially Compliant AV Control}
    \label{tab:algorithm}
    \begin{tabular}{cl}
        \toprule
        \textbf{Input:} & \(X_{\text{seq}}, \phi\) \\
        \textbf{Output:} & \(a_{\text{pred}}, \mathcal{L}\) \\
        \midrule
        1 & \(Z_{\text{seq}} \leftarrow \text{LSTM}(X_{\text{seq}})\) \\
          & Initialize sequence states using LSTM \\
        2 & \(Z_{\phi} \leftarrow \text{set}(\phi)\) \\
          & Set social preference parameter \\
        3 & \(a_{\text{pred}} \leftarrow \text{combine}(Z_{\text{seq}}, Z_{\phi})\) \\
          & Combine states to predict acceleration \\
        4 & \(\frac{\partial a_{\text{pred}}}{\partial x} \leftarrow \nabla a_{\text{pred}}\) \\
          & Compute gradient of predicted acceleration \\
        5 & \(\mathcal{L}_{\text{prediction}} = \int_{0}^{T} \left( a_{\text{pred}}(t) - a_{\text{true}}(t) \right)^2 \, dt\) \\
          & Define prediction loss \\

        6 & \(\mathcal{L}_{\text{cost}} = \int_{0}^{T} \left[ \cos(\phi) \cdot \tilde{U}_{\text{self}}(t) + \sin(\phi) \cdot \tilde{U}_{\text{collective}}(t) \right] \, dt\) \\
          & Define utility loss \\
        7 & \(\mathcal{L}_{\text{constraint}} = \mathcal{L}_{\text{smoothness}} + \mathcal{L}_{\text{trend}}\) \\
          & Define constraint loss \\
        8 & \(\mathcal{L}_{\text{smoothness}} = \int_{0}^{T} \left( \frac{d a_{\text{AV}}(t)}{dt} \right)^2 \, dt\) \\
           & Define smoothness component \\
        9 & \(\mathcal{L}_{\text{trend}} = \int_{0}^{T} \begin{array}{l} \left( U_{\text{self}}(t, \phi_1) - U_{\text{self}}(t, \phi_2) \right)^2 \\ + \left( U_{\text{collective}}(t, \phi_2) - U_{\text{collective}}(t, \phi_1) \right)^2 \end{array} \, dt\) \\
        
           & Define trend component \\
        10 & \(\mathcal{L} = \alpha \mathcal{L}_{\text{prediction}} + \beta \mathcal{L}_{\text{cost}} + \gamma \mathcal{L}_{\text{constraint}}\) \\
           & Compute total loss function \\
        11 & \(\nabla_{\Theta} \mathcal{L} \leftarrow \frac{\partial \mathcal{L}}{\partial \Theta}\) \\
           & Compute gradient of loss w.r.t. parameters \(\Theta\) \\
        12 & \(\Theta \gets \Theta - \eta \nabla_{\Theta} \mathcal{L}\) \\
           & Update parameters using learning rate \(\eta\) \\
        13 & \textbf{return} \(a_{\text{pred}}, \mathcal{L}\) \\
           & Return predicted acceleration and loss \\
        \bottomrule
    \end{tabular}
\end{table}

\subsection{Traffic Flow Simulation with Conventional CF Models}

The simulation of traffic flow involving the first, second, and third follower vehicles behind the AV is simulated using the Intelligent Driver Model (IDM), selected for its capacity to emulate naturalistic driving behaviors observed in human drivers~\cite{wang2023general, shang2023extending}. 
This choice is supported by the model's foundation in empirical data, which captures the variability and authenticity of human driving styles, as implemented in the computational framework. The IDM is valued for its ability to represent complex driving decisions---such as maintaining safe distances and responding to lead vehicle actions---rooted in established traffic flow theories, providing a dependable baseline for simulating the dynamics of follower entities within the traffic stream. The IDM is calibrated using real trajectory data specific to each vehicle to ensure that the simulated traffic flow responses reflect their distinct driving characteristics, operationalizing the abstract goal of modeling interdependent motion with physical realism. The acceleration dynamics of an IDM is governed by:
\begin{equation}
a_{\text{HV},i}(t) = a_{\text{max},i} \left[ 1 - \left( \frac{v_i(t)}{v_{0,i}} \right)^{\delta_i} - \left( \frac{s^*_i(t)}{s_i(t)} \right)^2 \right]
\end{equation}
where \(a_{\text{HV},i}(t)\) is the acceleration of the following HV, \(v_i(t)\) is its speed, \(v_{0,i}\) is the desired speed, \(\delta_i\) is the acceleration exponent, \(s_i(t)\) is the actual spacing, and \(s^*_i(t)\) is the desired spacing; this equation models acceleration as a function of speed and spacing, ensuring realistic driving behavior based on established principles, facilitating the simulation of follower dynamics in response to the controlled vehicle. The value of $s^*_i$ is computed as:
\begin{equation}
s^*_i(t) = s_{0,i} + v_i(t) \tau_i + \frac{v_i(t) \cdot (v_i(t) - v_{i-1}(t))}{2 \sqrt{a_{\text{max},i} \cdot b_i}}
\end{equation}
where \(s_{0,i}\) is the minimum spacing, \(v_i(t)\) is the speed of vehicle \(i\), \(\tau_i\) is the time gap, \(v_{i-1}(t)\) is the speed of the preceding vehicle, and \(b_i\) is the comfortable deceleration; this equation calculates the desired spacing to maintain safe and efficient following distances within the traffic flow, with the additional term accounting for the relative speed to ensure safe deceleration. The parameters for \(v_3\), \(v_4\), and \(v_5\) are calibrated using the IDM by optimizing against real-world trajectory data to minimize the spacing RMSE. This calibration process employs a grid search approach over the parameter space to identify the optimal values that best align the predicted spacing \(s_{\text{pred}}(t_i)\) with the actual spacing \(s_{\text{actual}}(t_i)\), defined as:
\begin{equation}
\text{RMSE}_{\text{spacing}} = \sqrt{\frac{1}{N} \sum_{i=1}^N (s_{\text{pred}}(t_i) - s_{\text{actual}}(t_i))^2}
\end{equation}
where \( N \) is the number of samples. The calibrated IDM parameters, derived through this optimization, are presented in Table~\ref{tab:idm_params} to ensure fidelity to real-world driving styles and enhance the accuracy of the simulated traffic flow. The resulting car-following trajectories, including speed and inter-vehicle spacing for a five-vehicle platoon, are visualized in Fig.~\ref{fig:idm_trajectories}, where the first and second vehicles use real data, and the third, fourth, and fifth vehicles use IDM with calibrated parameters to simulate car-following behavior.

\begin{figure}[t!]
    \centering
    \subfloat[Speed trajectories for a five-vehicle platoon]{\includegraphics[width=0.45\textwidth]{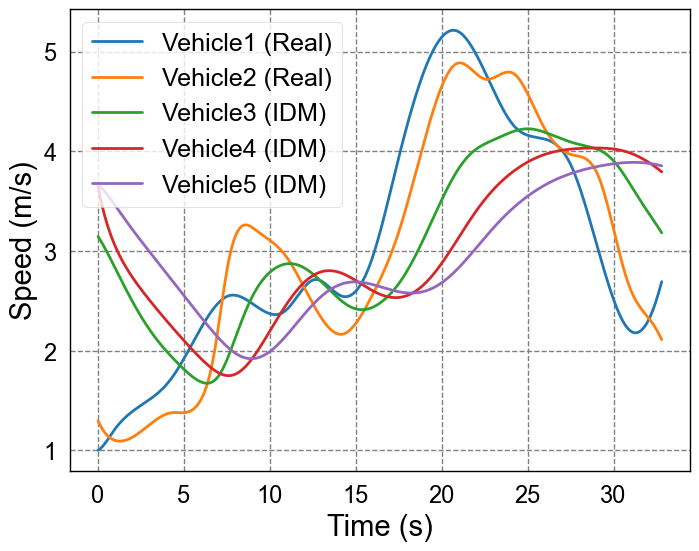}\label{fig:idm_speed_trajectory}} \hfill
    \subfloat[Inter-vehicle spacing for a five-vehicle platoon]{\includegraphics[width=0.45\textwidth]{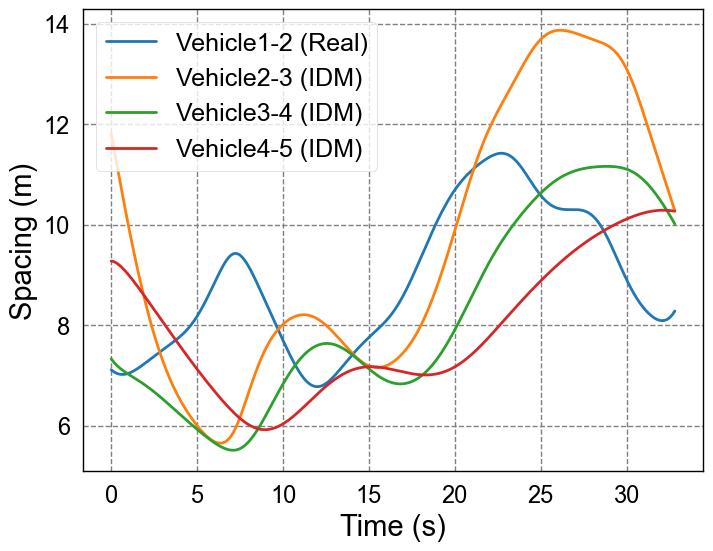}\label{fig:idm_spacing_trajectory}}
    \caption{Profiles of speed and inter-vehicle spacing for a five-vehicle platoon, with real data for the first and second vehicles and calibrated IDM for the third, fourth, and fifth vehicles, validating the accuracy of the IDM in replicating realistic car-following dynamics.}
    \label{fig:idm_trajectories}
\end{figure}

\begin{table}[t!]
    \centering
    \caption{Calibrated HV Parameters and Spacing RMSE of the IDM}
    \label{tab:idm_params}
    \begin{tabular}{c c c c c c c c}
        \toprule
        \multirow{2}{*}{\textbf{Vehicle}} & \multicolumn{6}{c}{\textbf{Parameters}} & \multirow{2}{*}{\textbf{RMSE (m)}} \\
        \cmidrule{2-7}
        & \(v_0\) & \(a_{\text{max}}\) & \(b\) & \(s_0\) & \(\tau\) & \(\delta\) & \\
        \midrule
        \(v_3\) & 5.00 & 1.19 & 0.50 & 3.00 & 1.50 & 2.81 & 3.95 \\
        \(v_4\) & 5.00 & 1.31 & 3.00 & 3.00 & 1.50 & 5.00 & 2.58 \\
        \(v_5\) & 5.00 & 1.70 & 0.50 & 3.00 & 1.50 & 5.00 & 0.35 \\
        \bottomrule
    \end{tabular}
\end{table}

This design ensures that the first, second, and third follower vehicles contribute realistically to the simulated traffic flow, providing a reasonable environment to study the impact of the controlled AV's learned driving behavior on the overall traffic dynamics.

\begin{figure*}[t!]
    \centering
    \subfloat[Speed profiles for \(\phi = 0\) (Egoistic)]{\includegraphics[width=0.3\textwidth]{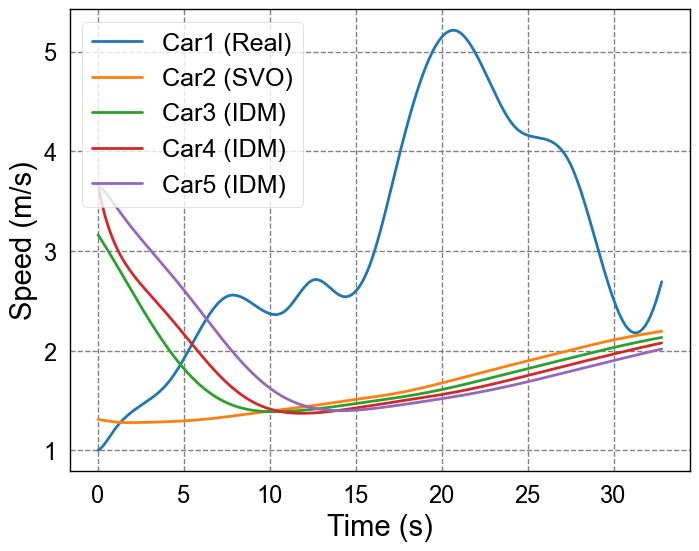}\label{fig:phi0_speed}} \hfill
    \subfloat[Speed profiles for \(\phi = \pi/4\) (Balanced)]{\includegraphics[width=0.3\textwidth]{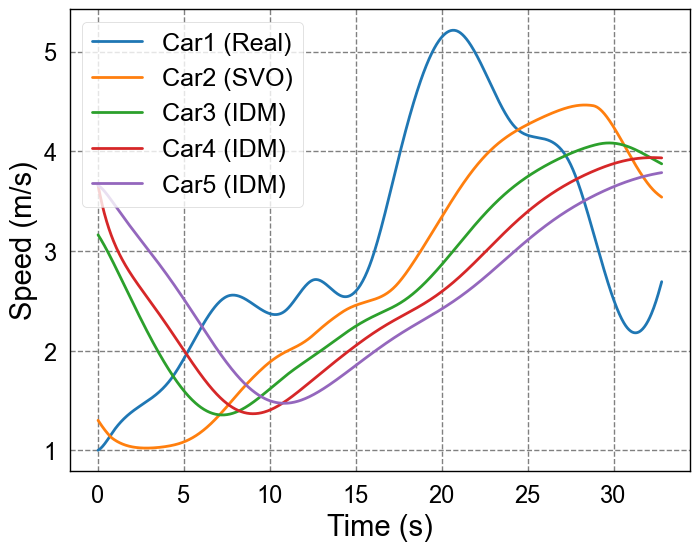}\label{fig:phi_pi_4_speed}} \hfill
    \subfloat[Speed profiles for \(\phi = \pi/2\) (Altruistic)]{\includegraphics[width=0.3\textwidth]{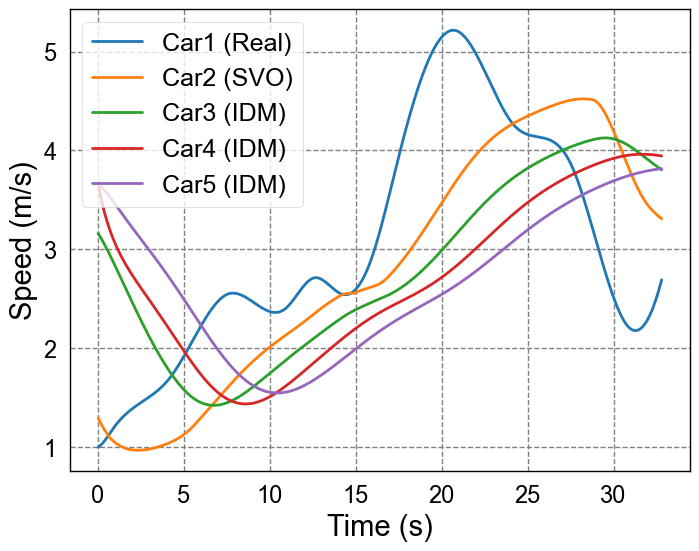}\label{fig:phi_pi_2_speed}}
    \vfill
    \subfloat[Spacing profiles for \(\phi = 0\) (Egoistic)]{\includegraphics[width=0.3\textwidth]{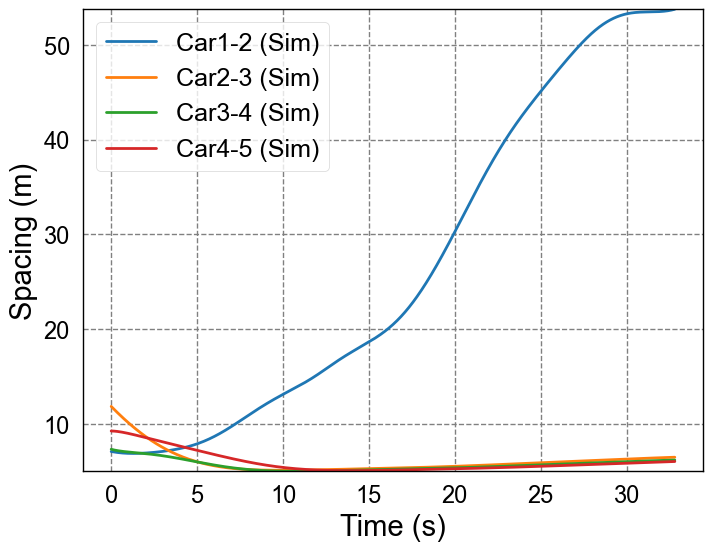}\label{fig:phi0_spacing}} \hfill
    \subfloat[Spacing profiles for \(\phi = \pi/4\) (Balanced)]{\includegraphics[width=0.3\textwidth]{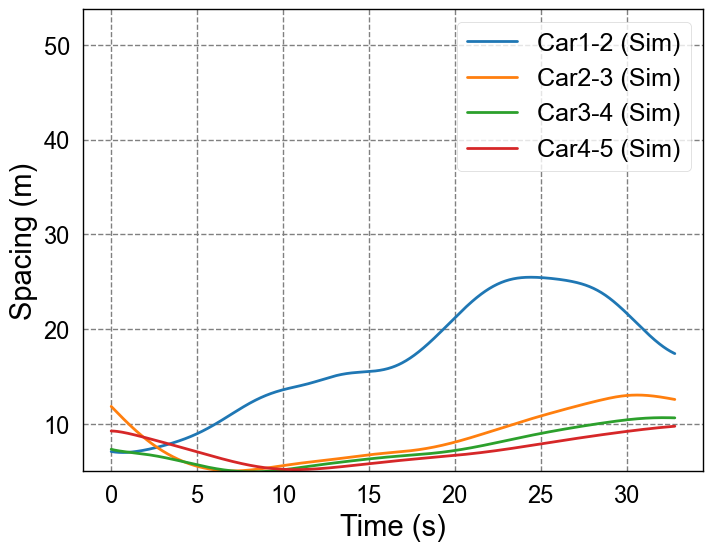}\label{fig:phi_pi_4_spacing}} \hfill
    \subfloat[Spacing profiles for \(\phi = \pi/2\) (Altruistic)]{\includegraphics[width=0.3\textwidth]{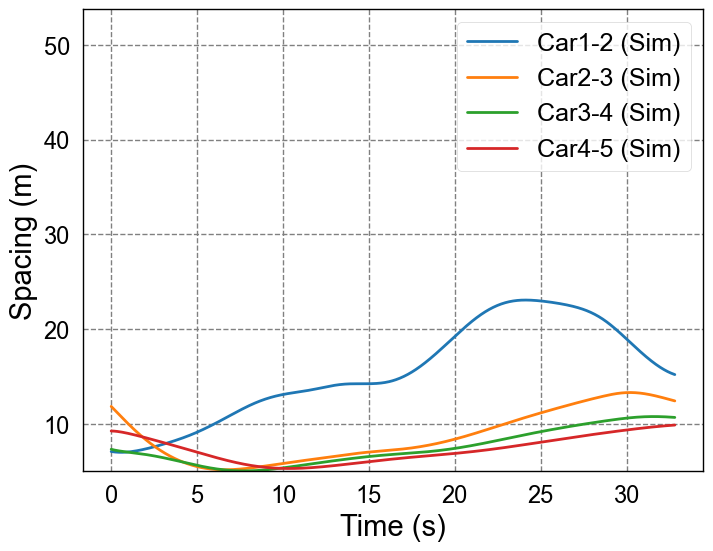}\label{fig:phi_pi_2_spacing}}
    \caption{Profiles of speed and spacing under different \(\phi\) values, illustrating the impact of social preference settings on traffic flow dynamics. At \(\phi = 0\), \(v_2\) adopts a conservative strategy, lagging behind \(v_1\) dynamics (1.0-2.0 m/s), while \(\phi = \pi/4\) and \(\phi = \pi/2\) show improved matching and smoother spacing, reflecting a shift from individual to system optimization based on SVO styles: egoistic (\(\phi = 0\)), balanced (\(\phi = \pi/4\)), and altruistic (\(\phi = \pi/2\)). It is worth noting that the y-axes of all spacing subfigures have been adjusted to start from 0 (or very close to 0) to provide an accurate visual representation. This eliminates visual artifacts where non-zero y-axis origins could misleadingly suggest near-zero or overlapping spacings (i.e., potential collisions). All actual minimum spacings remain strictly positive (approximately 5.0--5.5 m across scenarios), confirming collision-free behavior in all simulations.}
    \label{fig:trajectory_analysis}
\end{figure*}

The simulations were implemented in a custom Python environment using NumPy for numerical integration and Matplotlib for visualization, without relying on external traffic simulators such as CARLA, SUMO, or VISSIM. The hybrid closed-loop setup is as follows: (1) the lead vehicle $v_1$ strictly follows the real-world velocity profile extracted from the ARED dataset, providing a realistic and fixed reference trajectory; (2) the controlled AV $v_2$ computes its acceleration at each time step using the trained LSTM model, conditioned on the current spacing $s_{1,2}$, relative speed $\Delta v_{1,2}$, own speed $v_2$, and social preference $\phi$; (3) the following vehicles $v_3$, $v_4$, $v_5$ are modeled using calibrated IDM, with vehicle-specific parameters fitted to the corresponding ARED trajectories (see Table~\ref{tab:idm_params}); (4) all vehicle states (position, speed, spacing) are updated via forward Euler integration with a fixed time step $\Delta t = 1/30$ s, as:
\begin{align}
v_i(t + \Delta t) &= v_i(t) + a_i(t) \Delta t, \\
s_i(t + \Delta t) &= s_i(t) + [v_{i-1}(t) - v_i(t)] \Delta t.
\end{align}
where $a_i(t)$ is the acceleration from either the LSTM (for $v_2$) or IDM (for $v_3$--$v_5$). This approach enables precise control of the AV's socially aware decisions, direct preservation of the empirical lead behavior from ARED, and clear propagation of effects to the IDM followers in a closed-loop manner. The custom implementation was chosen to ensure exact replication of the ARED lead trajectory and straightforward comparison with real platoon observations.

\section{Numerical Results}\label{sec:analysis}
This section first evaluates Learn2Drive in the original 5-vehicle platoon, which serves as the core proof-of-concept. We then quantify the resulting self--collective trade-off using speed and an acceleration-based effort indicator, and finally examine two longer platoon configurations to assess whether the same qualitative behavior persists beyond the baseline setting.

The profiles of speed and spacing for the vehicular platoon under different values of \(\phi\) are illustrated in Fig.~\ref{fig:trajectory_analysis}, reflecting the adaptive AV control strategy's effect on traffic flow. The design incorporates three typical values of \(\phi\), i.e., 0, \(\pi/4\), and \(\pi/2\), to characterize \(v_2\) control objectives: \(\phi = 0\) prioritizes minimizing \(v_2\) energy consumption, while increasing \(\phi\) values shift the focus toward enabling follower vehicles (\(v_3, v_4, v_5\)) to achieve target speeds, thus enhancing overall system efficiency. This approach is validated by comparing real vehicle data (\(v_1\)) with simulated outcomes from the SVO and IDM. Speed profiles analysis reveals the significant influence of \(v_2\) control strategy on overall traffic flow. At \(\phi = 0\), \(v_2\) employs a conservative approach focused on minimizing its energy consumption, resulting in a lagged speed profiles that struggles to track \(v_1\) dynamics, maintaining low speeds (1.0--2.0~m/s) and constraining follower performance.

When \(\phi = 0\) (egoistic mode), the controlled AV ($v_2$) adopts a highly conservative strategy, resulting in speeds of only 1--2 m/s during periods of low lead vehicle velocity in the ARED dataset. This low speed is not unrealistic for the specific congested or stop-and-go conditions captured in the ARED ring experiment, where vehicles frequently operate at reduced speeds due to phantom traffic jams or downstream disturbances. The egoistic setting prioritizes minimizing the AV's own energy consumption (acceleration effort), leading to larger spacing and delayed response to the lead vehicle, which is consistent with the intended self-focused behavior under SVO theory. These results highlight the trade-off: strong egoistic preference can lead to conservative following and lower speeds in dense traffic, while higher \(\phi\) values promote more proactive and stable platoon behavior. All speeds remain within the physical range observed in the real ARED data. As \(\phi\) increases to \(\pi/4\), \(v_2\) adopts a more proactive acceleration strategy, particularly between 15--25 seconds, aligning closely with \(v_1\) speed and providing a better reference for followers. At \(\phi = \pi/2\), the strategy further balances individual and system goals, though marginal speed gains diminish, indicating a transition from local to global optimization---a core principle of the traffic flow control design. Spacing analysis further validates the systemic impact of \(v_2\) control strategy. At \(\phi = 0\), the inter-vehicle distance between \(v_1\) and \(v_2\) exhibits sharp fluctuations, escalating from an initial 12 meters to a peak of 53 meters, destabilizing the platoon. With \(\phi\) increasing to \(\pi/4\) and \(\pi/2\), the maximum distance remains similar, but fluctuations are notably smoothed, enhancing platoon stability and demonstrating \(v_2\) role in creating a more consistent traffic environment.

We note that the speed and spacing profiles for \(\phi = \pi/4\) (balanced/prosocial) and \(\phi = \pi/2\) (altruistic) are highly similar, with only marginal differences in upstream stabilization. This behavior is expected under the SVO framework: as \(\phi\) increases beyond \(\pi/4\), the collective utility weight \(\sin(\phi)\) approaches its maximum (1 at \(\pi/2\)), while the self-utility weight \(\cos(\phi)\) continues to decrease, but the marginal benefit to follower stability diminishes due to saturation of the collective objective. In other words, once the AV sufficiently prioritizes collective welfare, further increases in altruism yield limited additional improvements in the platoon setting. The choice of \(\phi = \pi/4\) and \(\phi = \pi/2\) represents typical intermediate and extreme altruistic preferences in SVO theory. To further illustrate this saturation effect, future work could include additional intermediate values (e.g., \(\phi = \pi/6\) or \(\phi = \pi/3\)) to map the full transition curve. All results remain consistent with the intended SVO trade-off and the ARED-based platoon dynamics.

\begin{figure*}[t!]
    \centering
    \subfloat[Energy consumption indicator of each vehicle, which quantifies acceleration-based energy cost and does not use traditional units like joules (J)]{\includegraphics[width=0.45\textwidth]{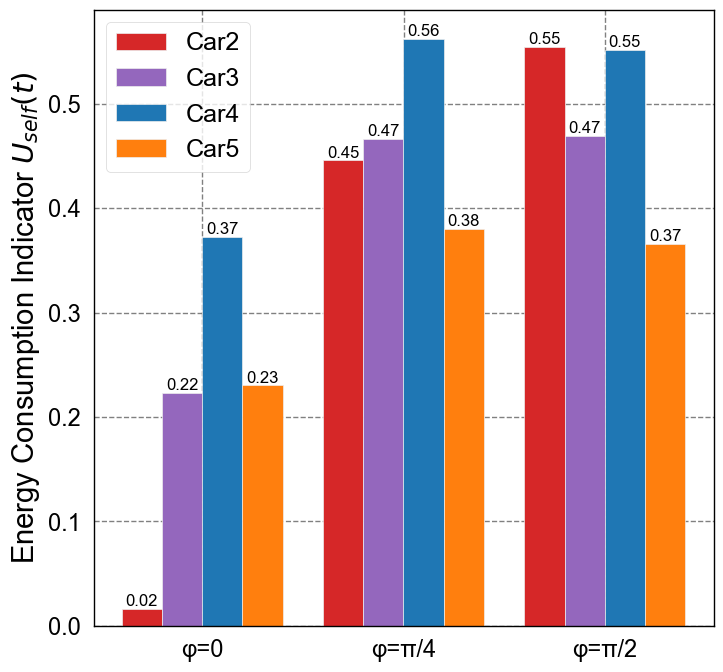}\label{fig:energy_consumption}}
    \hfill
    \subfloat[Average speed of each vehicle]{\includegraphics[width=0.45\textwidth]{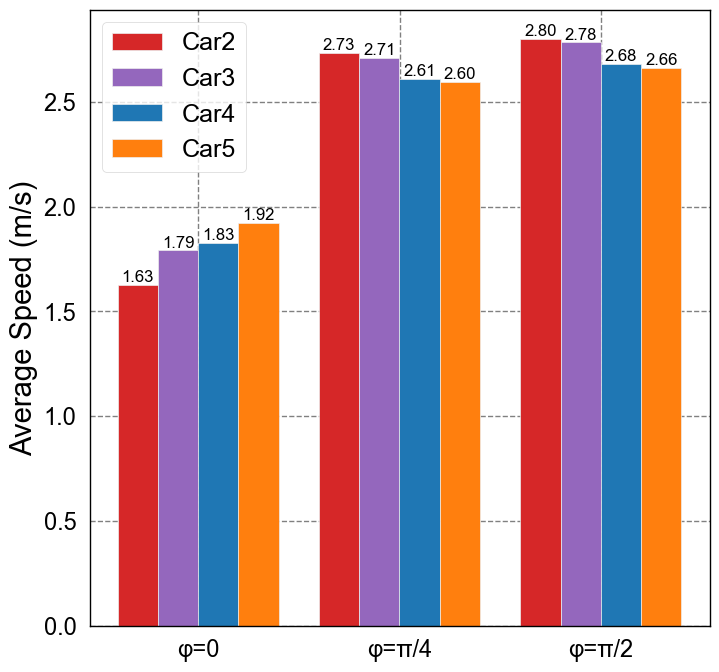}\label{fig:average_speeds}}   
    \caption{Energy consumption and average speed under different \(\phi\) values, highlighting the trade-offs in the Learn2Drive framework. \(v_2\) increased energy cost at \(\phi = \pi/4\) and \(\phi = \pi/2\) correlates with enhanced speed gains for followers, reflecting SVO styles: egoistic (\(\phi = 0\)), balanced (\(\phi = \pi/4\)), and altruistic (\(\phi = \pi/2\)).}
    \label{fig:performance_metrics}
\end{figure*}

Performance metrics, including energy consumption and average speed, are presented in Fig.~\ref{fig:performance_metrics}, offering a quantitative evaluation of the framework's trade-offs under varying \(\phi\) values. Energy consumption analysis of \(v_2\), the key executor of traffic flow control, reveals a trade-off with system efficiency, where the energy consumption indicator quantifies acceleration-based energy cost and does not use traditional units like joules (J). At \(\phi = 0\), \(v_2\) energy use is minimal at 0.02, aligning with its self-focused goal. At \(\phi = \pi/4\), it rises to 0.45 (a 2621.81\% increase), and at \(\phi = \pi/2\), it reaches 0.55 (a 3284.22\% increase), indicating a significant energy cost to enhance system performance, with diminishing marginal gains at higher \(\phi\) values, as shown in Table~\ref{tab:energy_consumption}. The ``performance transfer effect'' from \(v_2\) strategy demonstrates a performance transfer effect for follower vehicles (\(v_3, v_4, v_5\)). Energy consumption for \(v_3\) increases from 0.22 at \(\phi = 0\) to 0.47 at \(\phi = \pi/4\) (109.01\%) and stabilizes, while \(v_4\) and \(v_5\) show moderate peaks at \(\phi = \pi/4\) (0.56 and 0.38, respectively) with slight declines at \(\phi = \pi/2\). Moreover, average speed improvements of each vehicle are notable: \(v_3\) rises from 1.79 m/s to 2.71 m/s (51.40\%), \(v_4\) from 1.83 m/s to 2.61 m/s (42.99\%), and \(v_5\) from 1.92 m/s to 2.66 m/s (34.92\%), as summarized in Table~\ref{tab:average_speed}, with the effect diminishing with distance from \(v_2\).

One of the primary contributions of this study lies in validating the feasibility of optimizing traffic flow through socially compliant control of AVs, e.g., \(v_2\). The transition from \(\phi = 0\) to \(\phi = \pi/2\) reflects a shift from individual to system optimization. At \(\phi = 0\), \(v_2\) self-interested strategy creates a performance bottleneck, limiting follower potential. As \(\phi\) increases, \(v_2\) proactive behavior enhances follower conditions (\(v_3, v_4, v_5\)), improving system efficiency. This hierarchical control effect, with stronger influence on proximal vehicles (e.g., \(v_3\)) and diminishing impact distally (e.g., \(v_5\)), suggests that strategic placement of intelligent vehicles can efficiently regulate traffic. The energy-efficiency trade-off is evident, with \(\phi = \pi/4\) offering an optimal balance where \(v_2\) energy increase yields significant speed gains for followers, while \(\phi = \pi/2\) shows diminishing returns. This indicates an optimal control range, beyond which additional effort may reduce cost-effectiveness. The hierarchical nature of \(v_2\) influence provides insights for large-scale traffic control, advocating for targeted deployment of intelligent vehicles to enhance system-wide performance with manageable complexity.

Another significant contribution is the comprehensive analysis of a single AV’s (\(v_2\)) impact on the collective traffic flow by treating the three following vehicles (\(v_3, v_4, v_5\)) as a unified entity representing the broader traffic system. By modeling these followers as a collective, the study quantifies how \(v_2\)’s socially compliant behavior, driven by varying \(\phi\), influences overall traffic dynamics. At lower \(\phi\) values (e.g., \(\phi = 0\)), \(v_2\)’s self-interested behavior saves its own energy consumption but leads to performance bottlenecks, such as increased speed deviations and inconsistent spacing across the collective, contributing to traffic instability and the propagation of traffic waves. In contrast, at higher \(\phi\) values (e.g., \(\phi = \pi/4\) and \(\phi = \pi/2\)), \(v_2\)’s proactive adjustments reduce speed deviations, improve spacing consistency, and mitigate traffic wave propagation, leading to smoother traffic dynamics and reduced congestion across the collective. This regulation is achieved without requiring vehicle connectivity, as \(v_2\) relies solely on local observations, such as spacing and relative speeds, to make socially compliant decisions, demonstrating the framework’s robustness and implementability in diverse traffic scenarios. Furthermore, the study reveals the potential scalability of this approach, suggesting that a single AV’s influence on the collective can extend to larger traffic networks by strategically modulating its behavior to dampen instabilities and enhance system-wide efficiency. This insight underscores the Learn2Drive framework’s capability to serve as a scalable and practical solution for optimizing mixed traffic environments through targeted AV deployment.

\begin{table}[t!]
    \centering
    \caption{Percentage Change in Energy Consumption Indicator Across \(\phi\) Values (\(\phi = 0\): Egoistic, \(\phi = \pi/4\): Balanced, \(\phi = \pi/2\): Altruistic) Relative to Egoistic Case (\(\phi = 0\))}
    \label{tab:energy_consumption}
    \begin{tabular}{cccc}
        \toprule
        Vehicle & \(\phi = 0\) & \(\phi = \pi/4\) & \(\phi = \pi/2\) \\
        \midrule
        \( v_2 \) & -- & +2621.81\% & +3284.22\% \\
        \( v_3 \) & -- & +109.01\% & +110.39\% \\
        \( v_4 \) & -- & +50.89\% & +48.18\% \\
        \( v_5 \) & -- & +65.06\% & +58.99\% \\
        \bottomrule
    \end{tabular}
\end{table}

\begin{table}[t!]
    \centering
    \caption{Percentage Change in Average Speed (m/s) Across \(\phi\) Values (\(\phi = 0\): Egoistic, \(\phi = \pi/4\): Balanced, \(\phi = \pi/2\): Altruistic) Relative to Egoistic Case (\(\phi = 0\)) }
    \label{tab:average_speed}
    \begin{tabular}{cccc}
        \toprule
        Vehicle & \(\phi = 0\) & \(\phi = \pi/4\) & \(\phi = \pi/2\) \\
        \midrule
        \( v_2 \) & -- & +68.03\% & +72.17\% \\
        \( v_3 \) & -- & +51.40\% & +55.42\% \\
        \( v_4 \) & -- & +42.99\% & +46.86\% \\
        \( v_5 \) & -- & +34.92\% & +38.39\% \\
        \bottomrule
    \end{tabular}
\end{table}

\begin{figure*}[t]
    \centering

    \begin{subfigure}[t]{0.24\textwidth}
        \centering
        \includegraphics[width=\linewidth]{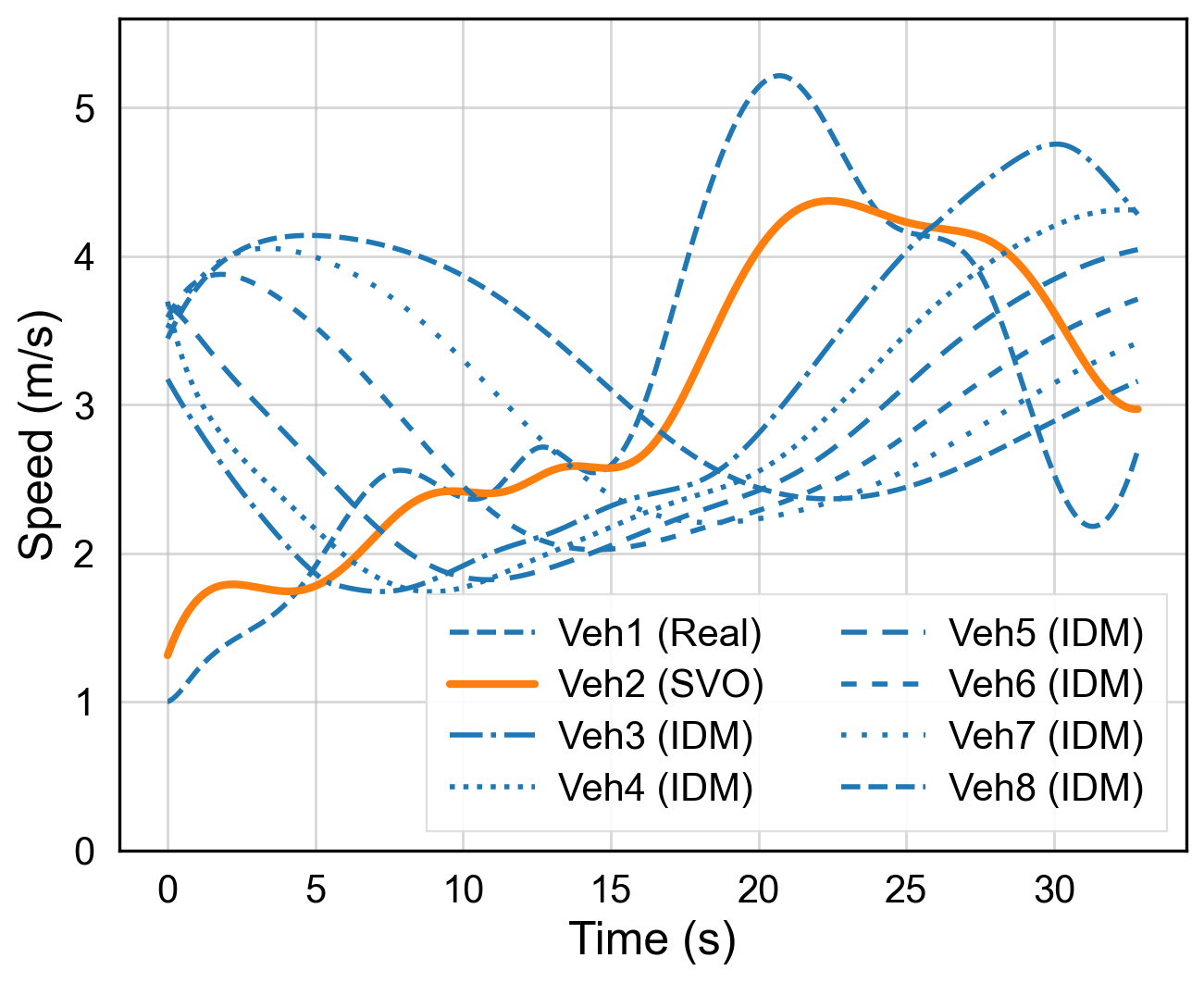}
        {\small 8-vehicle, $\phi=\pi/4$ — Speed}
        \label{fig:8veh_phi45_speed}
    \end{subfigure}
    \hfill
    \begin{subfigure}[t]{0.24\textwidth}
        \centering
        \includegraphics[width=\linewidth]{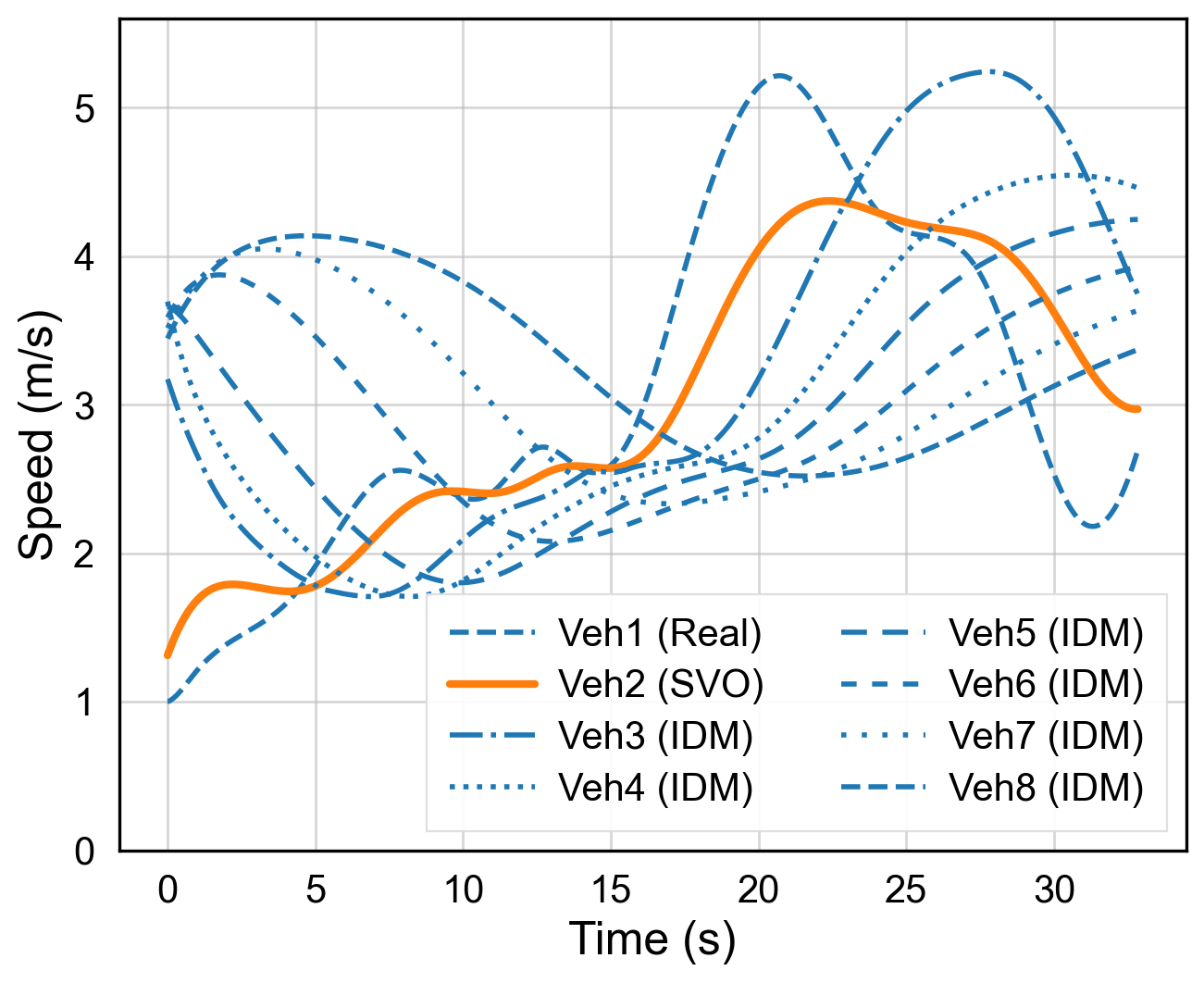}
        {\small 8-vehicle, $\phi=\pi/2$ — Speed}
        \label{fig:8veh_phi90_speed}
    \end{subfigure}
    \hfill
    \begin{subfigure}[t]{0.24\textwidth}
        \centering
        \includegraphics[width=\linewidth]{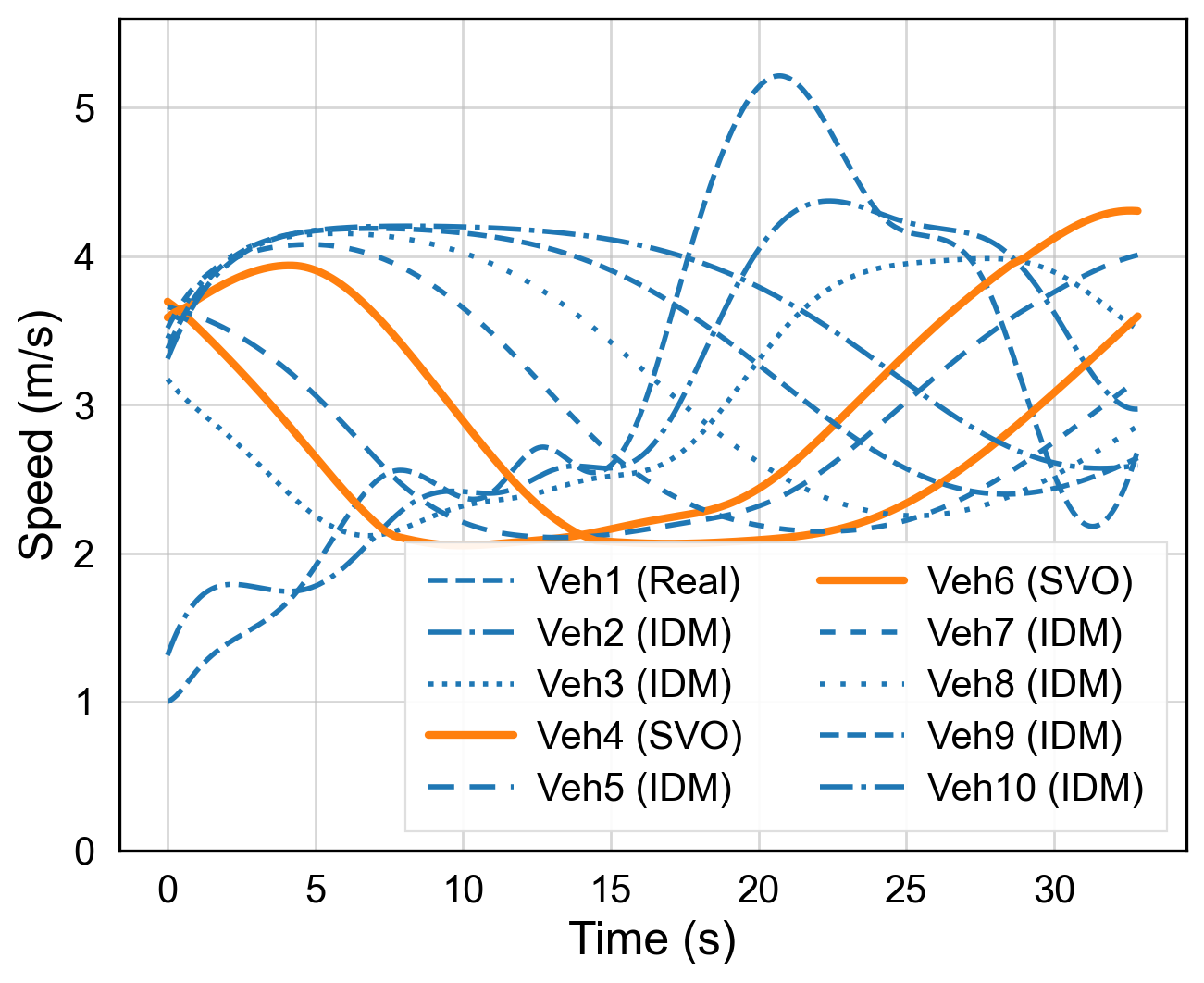}
        {\small 10-vehicle, $\phi=\pi/4$ — Speed}
        \label{fig:10veh_phi45_speed}
    \end{subfigure}
    \hfill
    \begin{subfigure}[t]{0.24\textwidth}
        \centering
        \includegraphics[width=\linewidth]{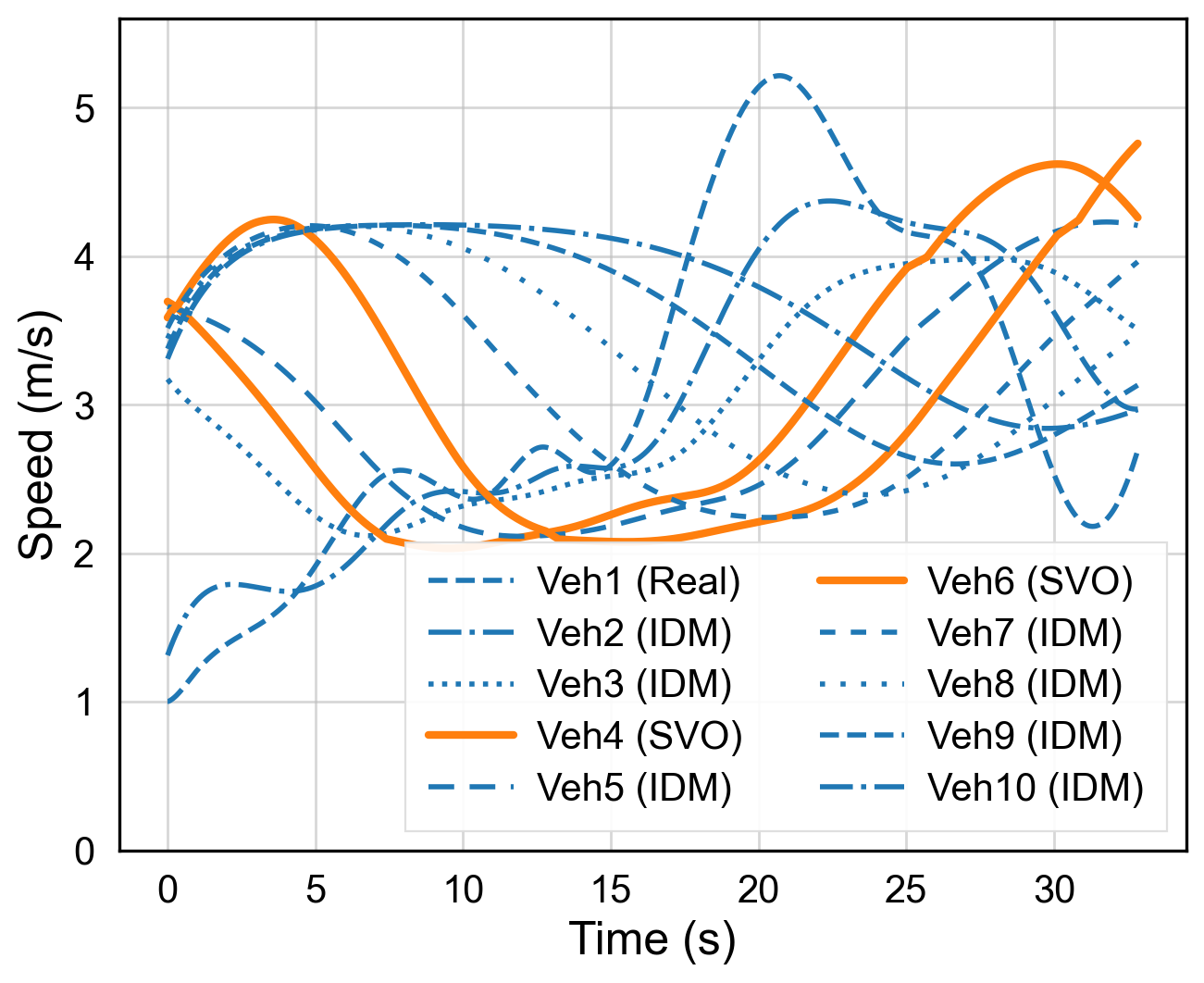}
        {\small 10-vehicle, $\phi=\pi/2$ — Speed}
        \label{fig:10veh_phi90_speed}
    \end{subfigure}

    \vspace{0.6em} 

    \begin{subfigure}[t]{0.24\textwidth}
        \centering
        \includegraphics[width=\linewidth]{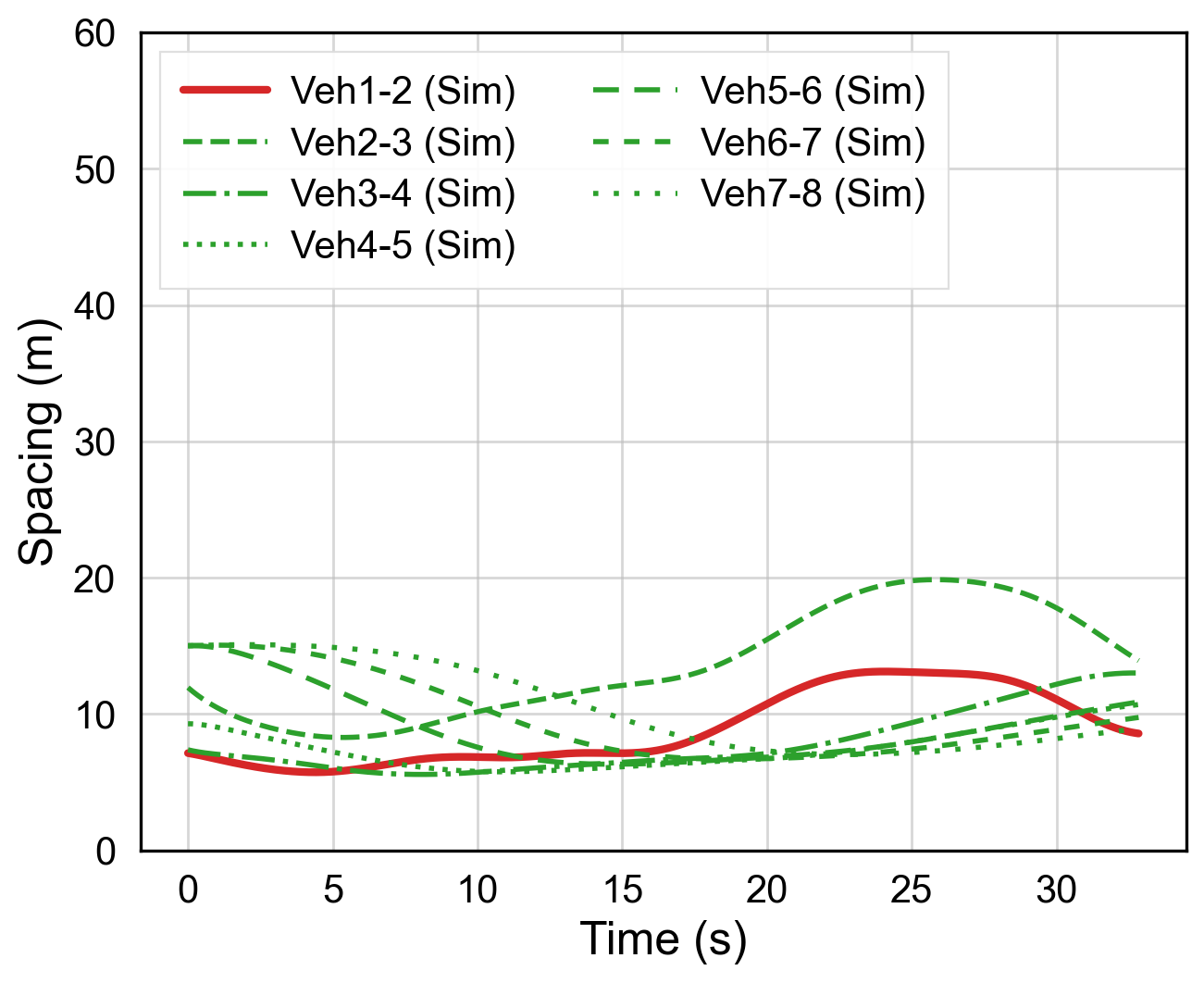}
        {\small 8-vehicle, $\phi=\pi/4$ — Spacing}
        \label{fig:8veh_phi45_spacing}
    \end{subfigure}
    \hfill
    \begin{subfigure}[t]{0.24\textwidth}
        \centering
        \includegraphics[width=\linewidth]{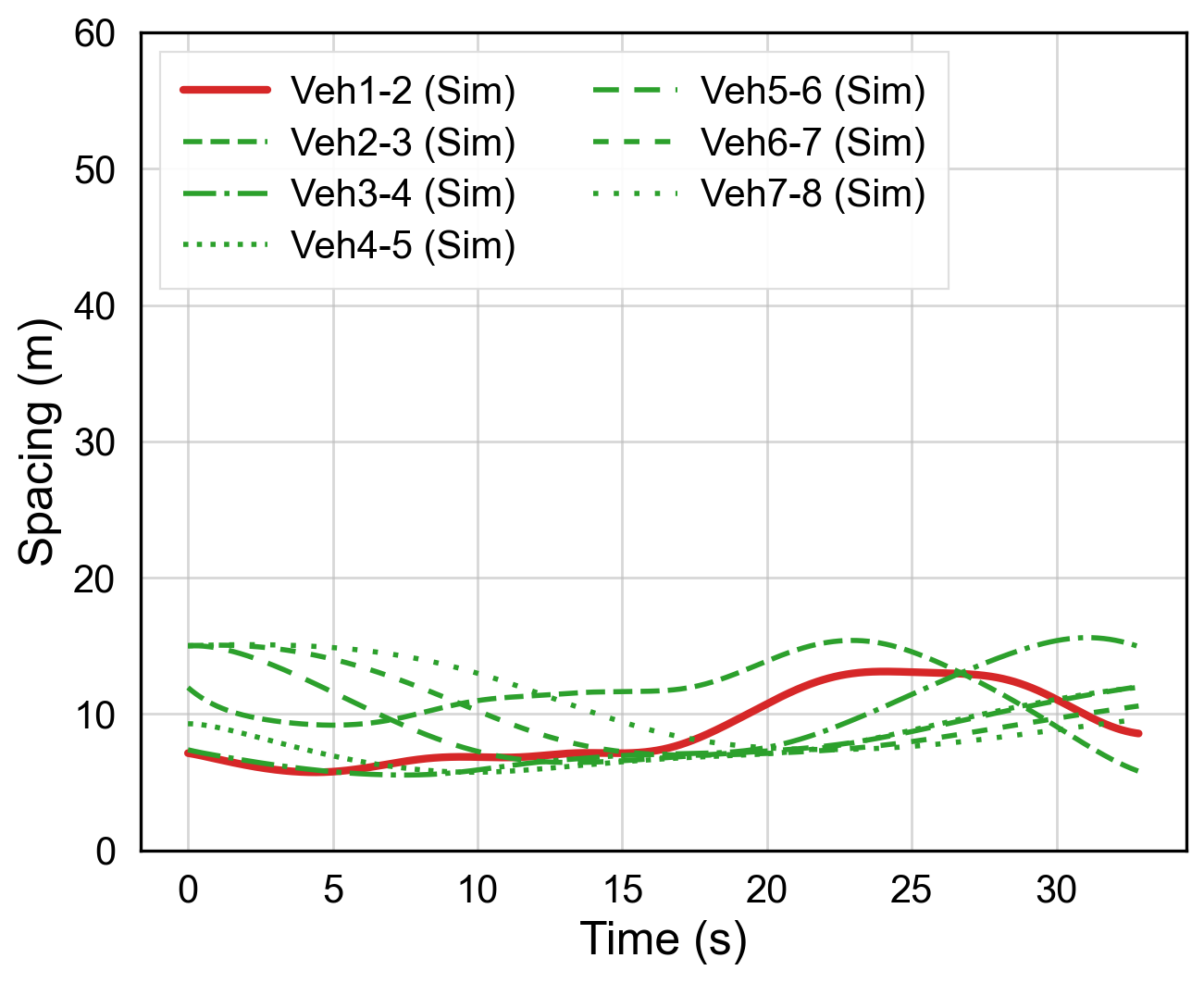}
        {\small 8-vehicle, $\phi=\pi/2$ — Spacing}
        \label{fig:8veh_phi90_spacing}
    \end{subfigure}
    \hfill
    \begin{subfigure}[t]{0.24\textwidth}
        \centering
        \includegraphics[width=\linewidth]{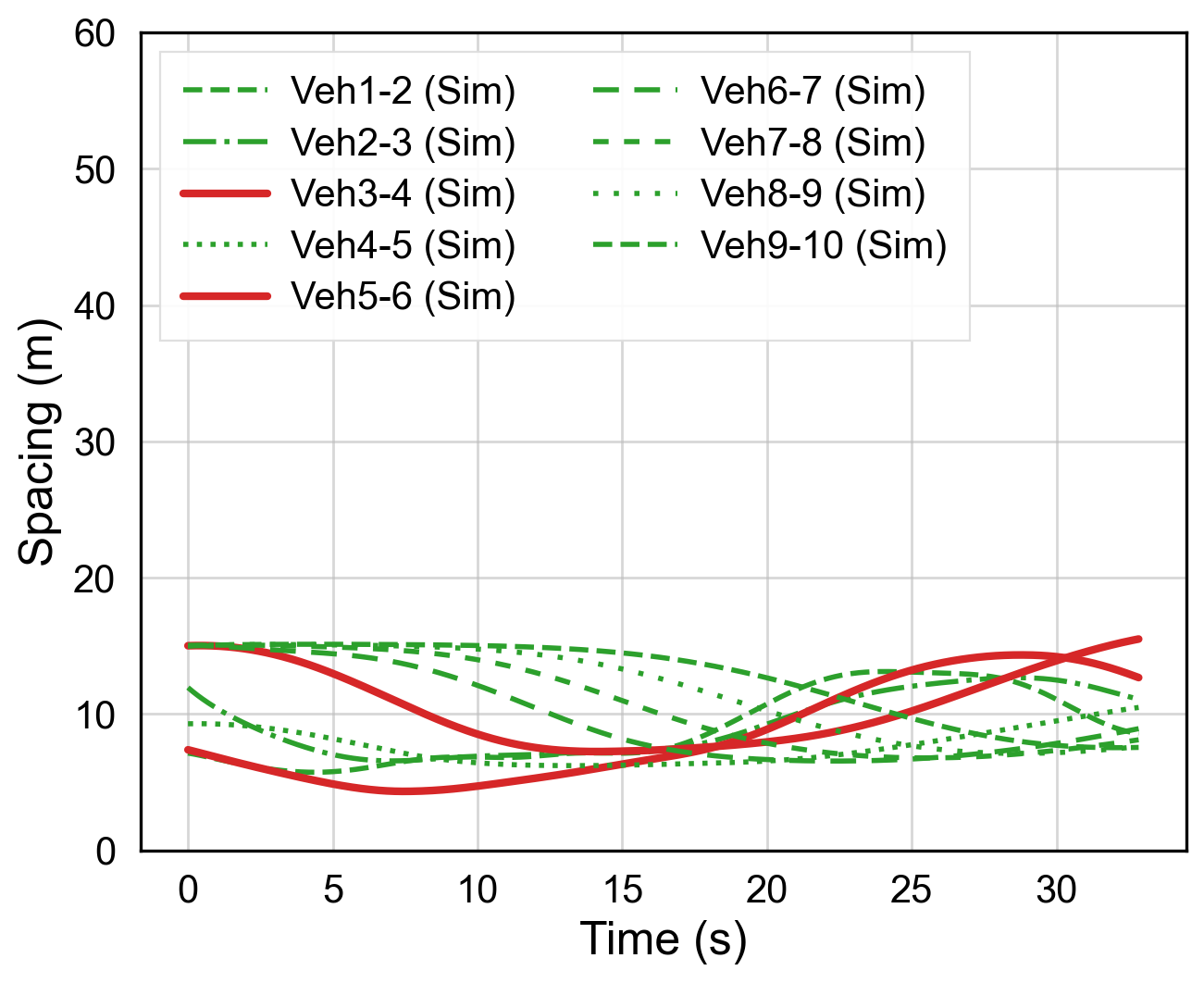}
        {\small 10-vehicle, $\phi=\pi/4$ — Spacing}
        \label{fig:10veh_phi45_spacing}
    \end{subfigure}
    \hfill
    \begin{subfigure}[t]{0.24\textwidth}
        \centering
        \includegraphics[width=\linewidth]{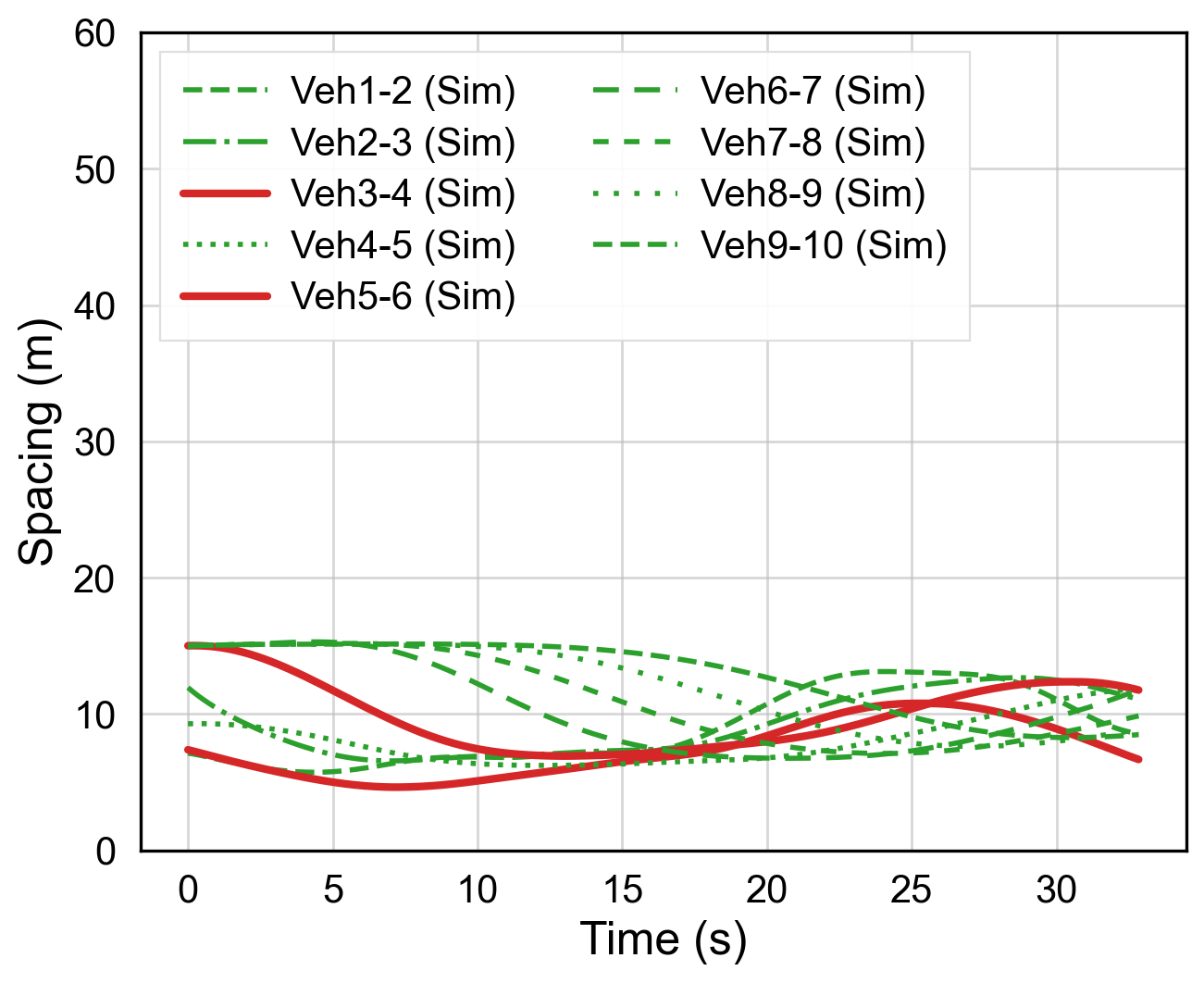}
        {\small 10-vehicle, $\phi=\pi/2$ — Spacing}
        \label{fig:10veh_phi90_spacing}
    \end{subfigure}

    \caption{Combined speed \& spacing profiles under SVO settings (2×4 grid).
    Top row: speed profiles for 8-vehicle ($\phi=\pi/4$, $\pi/2$) and 10-vehicle ($\phi=\pi/4$, $\pi/2$) configurations.
    Bottom row: spacing evolution for the corresponding configurations.
    In spacing plots, AV–front spacings are highlighted in red per our plotting convention.}
    \label{fig:profiles_all_2x4}
\end{figure*}

Having established the detailed performance trade-offs and hierarchical control effects in the baseline 5-vehicle scenario, we further expand our analysis to ensure the framework's robustness. To assess generality beyond the initial 5-vehicle platoon, two additional configurations are considered. These configurations enable evaluation of the SVO-guided control strategy in longer and more structurally complex mixed-traffic streams. For each configuration, the system is evaluated under $\phi \in \{0, \pi/4, \pi/2\}$, and the reported results include speed profiles, spacing evolution, an acceleration-based energy indicator, and a time-headway (THW) safety metric computed for each vehicle. Across both configurations, the qualitative patterns observed in the 5-vehicle case—namely, enhanced responsiveness and smoother flow at larger $\phi$—persist, with diminishing incremental improvements from $\phi=\pi/4$ to $\phi=\pi/2$. Consolidated comparison results across platoon sizes and SVO settings are provided in Fig.~\ref{fig:profiles_all_2x4}. The average THW from Fig.~\ref{fig:thw_all} exhibits consistent, interpretable trends across $\phi$. In the 5-vehicle case, the controlled AV (Veh2) shows a large THW under $\phi=0$ due to conservative following, which decreases as $\phi$ increases (reflecting more proactive, flow-oriented behavior), while the followers (Veh3–Veh5) maintain THW values clustered around 3s with mild reductions from $\phi=\pi/4$ to $\phi=\pi/2$. In the 8-vehicle case, most vehicles remain near 3s across $\phi$, with a localized increase for Veh3 at $\phi=\pi/4$ indicating a buffering effect upstream of the AV. In the 10-vehicle case with two AVs, THW values become more homogeneous as $\phi$ increases, suggesting complementary regulation between the two AVs that reduces variability for trailing vehicles. Overall, larger $\phi$ values yield more uniform and moderate THW distributions without compromising minimum spacing, consistent with the observed smoothing in speed/spacing and the saturation effect from $\phi=\pi/4$ to $\phi=\pi/2$. Future work will examine broader road geometries, multi-lane interactions, and varying AV penetration rates to further generalize these findings.

\begin{figure*}[t]
    \centering
    \begin{subfigure}[t]{0.31\textwidth}
        \centering
        \includegraphics[width=\linewidth]{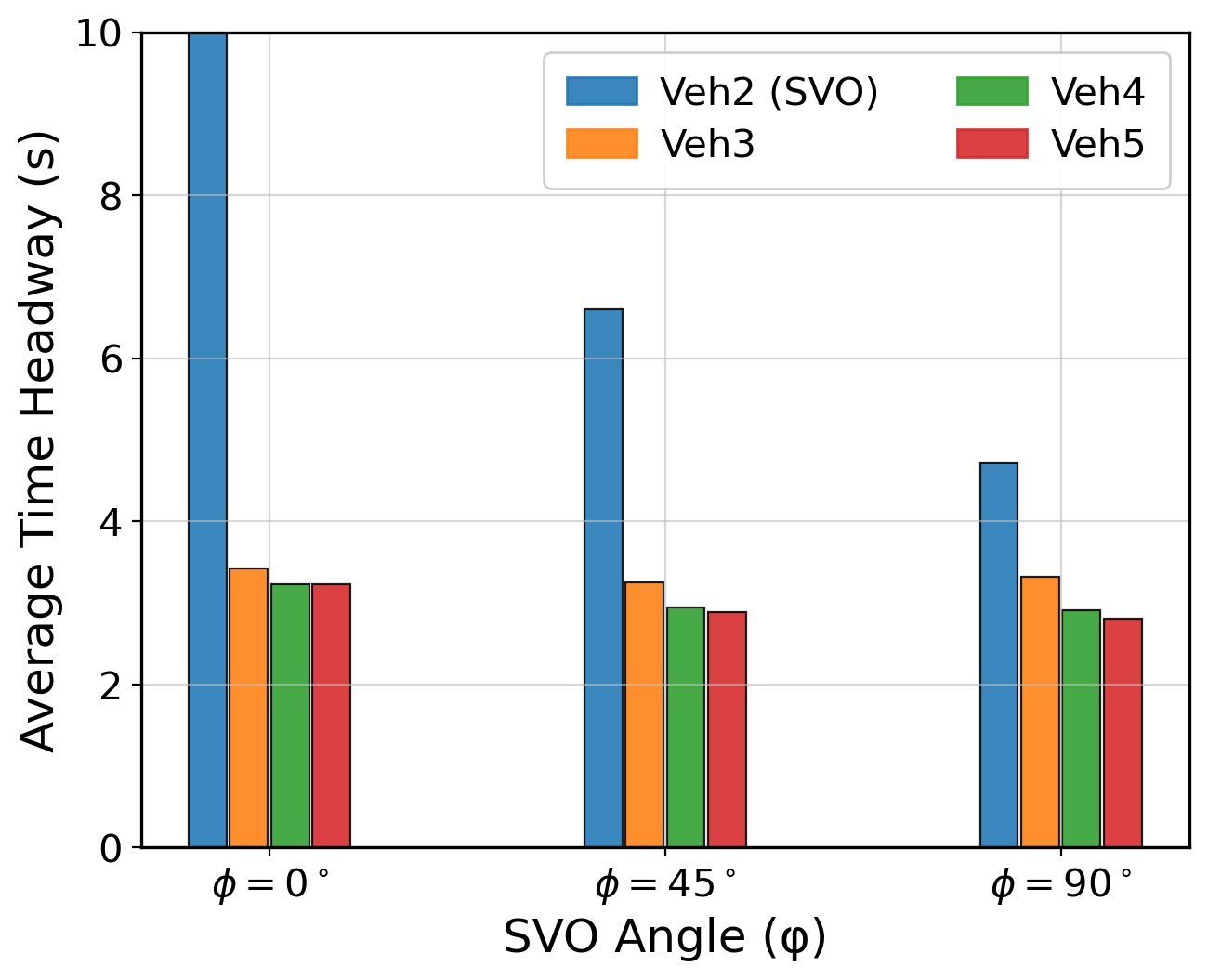}
        \caption{5-vehicle platoon}
        \label{fig:thw_5veh}
    \end{subfigure}
    \hfill
    \begin{subfigure}[t]{0.31\textwidth}
        \centering
        \includegraphics[width=\linewidth]{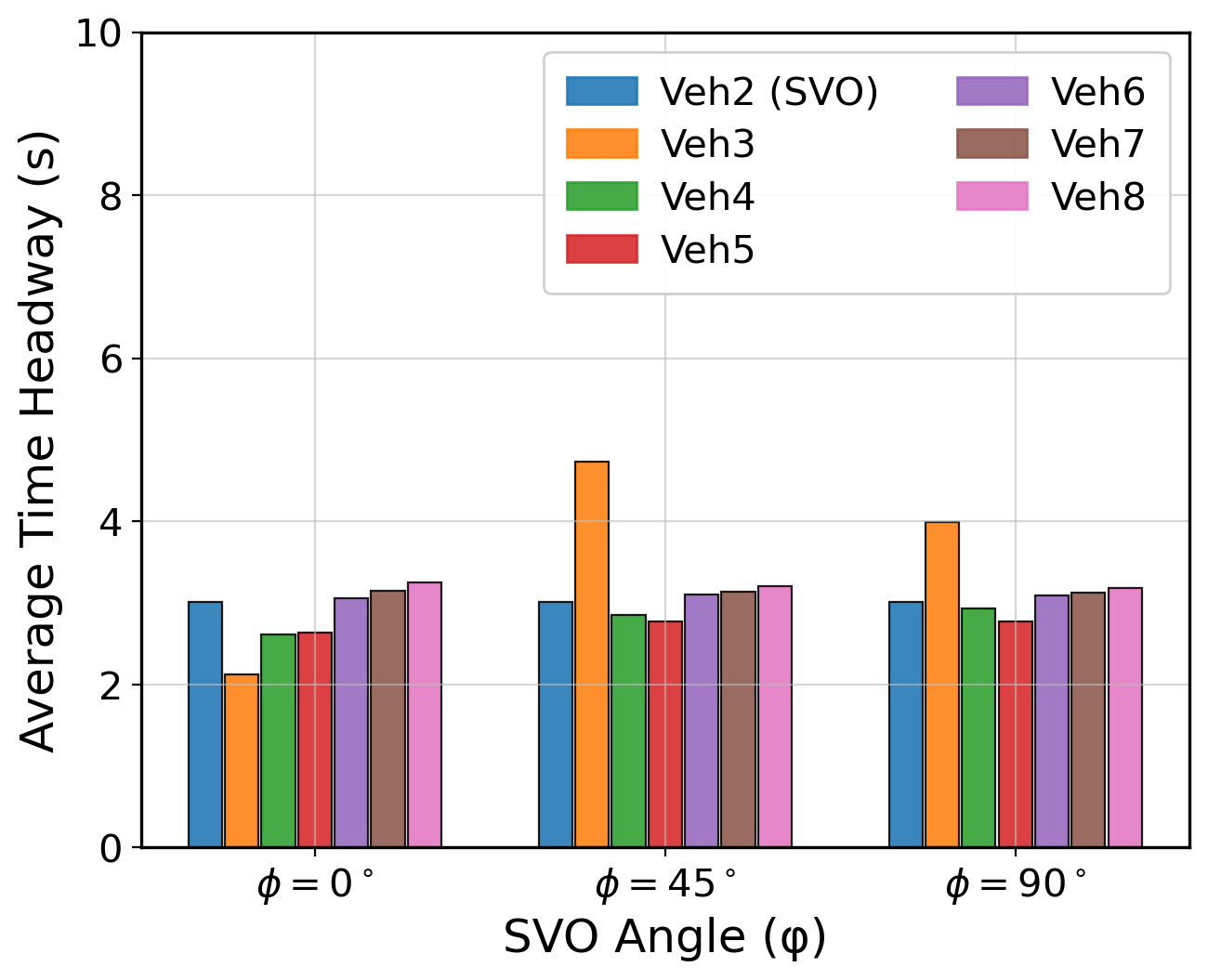}
        \caption{8-vehicle platoon}
        \label{fig:thw_8veh}
    \end{subfigure}
    \hfill
    \begin{subfigure}[t]{0.31\textwidth}
        \centering
        \includegraphics[width=\linewidth]{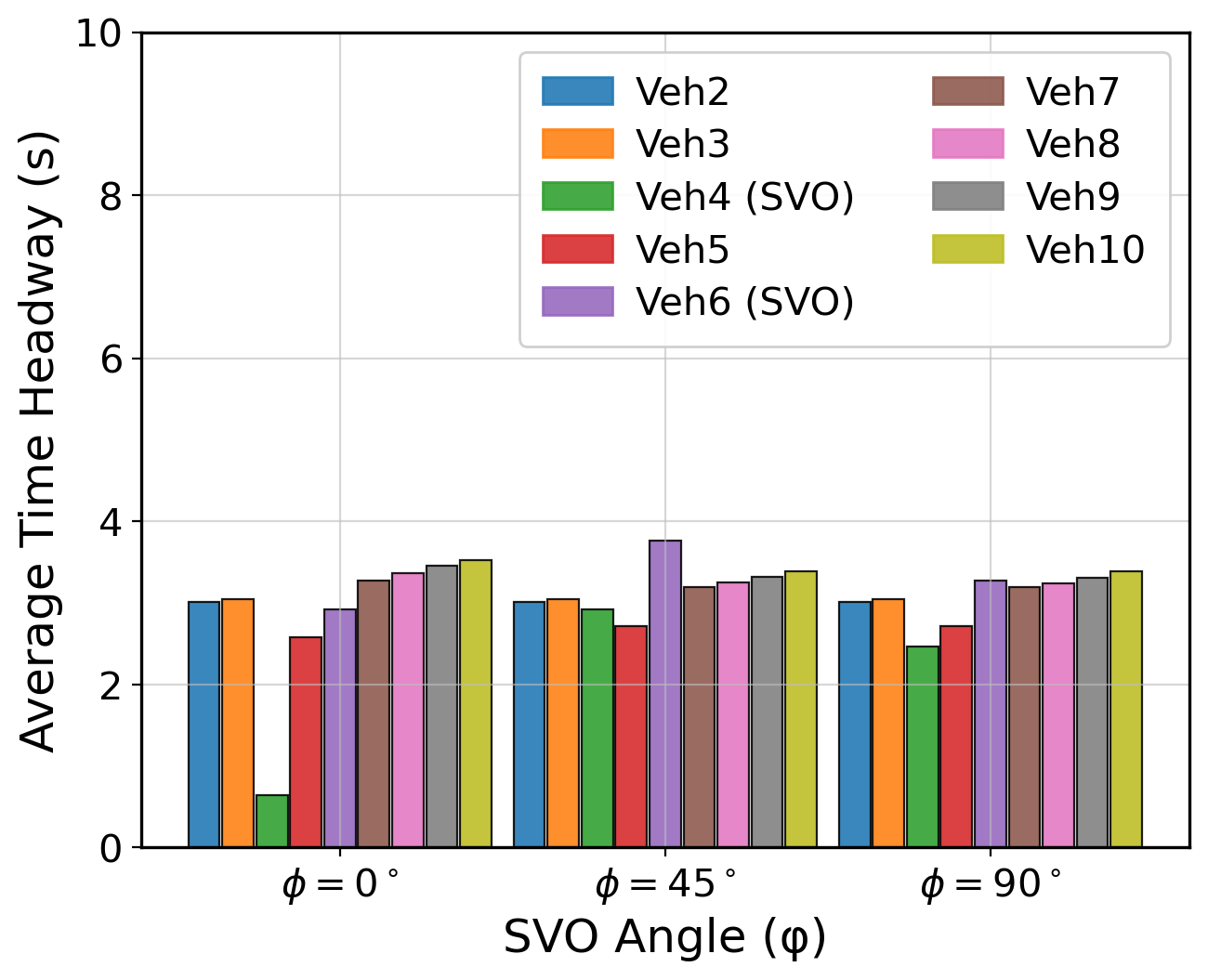}
        \caption{10-vehicle platoon}
        \label{fig:thw_10veh}
    \end{subfigure}

    \caption{Average time headway (THW) across SVO settings. Bars show mean THW (s) for each vehicle under $\phi \in \{0,\ \pi/4,\ \pi/2\}$.}
    \label{fig:thw_all}
\end{figure*}

\section{Discussion \& Conclusion}\label{sec:con}

This study has introduced the Learn2Drive framework, an advanced methodology for socially compliant AV control that integrates LSTM networks, physics-informed constraints, and the SVO principle to improve traffic dynamics. The primary contribution lies in the development of an adaptable control strategy that leverages the behavior of a single AV to influence broader traffic flow patterns. This is demonstrated through the AV's ability to adjust its acceleration profile in response to varying social preference settings, thereby optimizing energy consumption while enhancing overall traffic responsiveness. The integration of the IDM for follower vehicles, calibrated using empirical trajectory data, further supports the framework by simulating realistic human-driven responses. Initial results indicate successful state alignment and reduced performance discrepancies, reinforcing the potential of this approach to address a critical gap in current AV designs, which often prioritize individual efficiency over cooperative dynamics.

Nonetheless, the study has certain limitations that merit further investigation. The foundational evaluation was initially conducted in a specific 5-vehicle scenario (with the AV as the second vehicle). This setup was intentionally designed as a controlled proof-of-concept to cleanly isolate the direct impact of varying social preference settings on immediate following vehicles. We subsequently extended the framework to 8- and 10-vehicle configurations to demonstrate the generality of these SVO-guided behaviors, the scope remains focused on platoon-level dynamics. 
While the proposed Learn2Drive framework demonstrates promising performance in nominal mixed-traffic conditions, real-world ACC systems are subject to various disturbances and noise sources, including sensor measurement errors, environmental factors (e.g., weather, lighting), and unpredictable human driving behaviors (e.g., sudden braking or cut-ins). The current LSTM architecture may provide some tolerance under nominal perturbations through its long-term memory mechanism, which can filter out short-term fluctuations. During training, small Gaussian noise (standard deviation 0.005–0.01) is deliberately added to inputs and predicted accelerations to enhance resilience to measurement perturbations. However, the framework does not yet incorporate explicit mechanisms for handling severe or persistent disturbances, such as sensor failures, outliers, or extreme adversarial scenarios. Addressing these challenges through techniques like explicit noise modeling, Kalman filtering for state estimation, adversarial training, or integration of uncertainty quantification represents a critical direction for future work to improve the safety and reliability of socially compliant AV control in real-world deployment.

A critical consideration for real-world deployment of ACC and socially compliant AV control is resilience to adversarial attacks (e.g., sensor spoofing, command injection, or subtle policy manipulation), which can induce widespread traffic instability without obvious detection. Building on our prior study~\cite{li2022detecting,li2024detecting}, where we analyzed the microscopic and macroscopic impacts of subtle cyberattacks on ACC vehicles (malicious control manipulation, data poisoning, and DoS) and developed a GAN-based trajectory-level anomaly detection framework, the current Learn2Drive controller benefits from temporal-noise robustness inherent to LSTMs and from measurement disturbances present in ARED. We additionally inject small Gaussian noise (std $\approx 0.005$--$0.01$) into inputs and predictions during training to improve generalization to measurement perturbations. Nevertheless, this work focuses on nominal mixed-traffic behavior and does not incorporate explicit defenses against targeted adversarial perturbations (e.g., gradient-based attacks on the LSTM or policy-level manipulation). Addressing such threats—via robust/min–max optimization, adversarial training, uncertainty quantification, or the integration of online attack-detection modules as explored in~\cite{li2024detecting,wang2023novel,wang2025analytical,wang2023optimal}—is an important direction for future research to ensure safety-critical performance and resilience in deployment.

Future research will focus on extending the framework to encompass larger and more heterogeneous traffic environments, refining calibration using a broader set of datasets, and incorporating adaptive \(\phi\) value adjustments based on perceived information, such as traffic density and surrounding vehicle behavior, to enhance social compliance. 
Additionally, the high similarity between results at \(\phi = \pi/4\) and \(\phi = \pi/2\) reflects the diminishing marginal returns of further increasing altruism beyond prosocial levels in this platoon setting. Future work could explore a denser sampling of \(\phi\) values (e.g., \(\phi = \pi/6\), \(\pi/3\)) to better characterize the transition and saturation effects in the SVO control design. Moreover, exploring a wider range of evaluation metrics will strengthen the assessment of the framework’s performance. Incorporating live traffic data and dynamic system feedback will also be investigated to develop a more flexible and scalable control architecture, facilitating the broader adoption of socially compliant AVs.

\clearpage
\newpage
\bibliographystyle{IEEEtran}

\bibliography{refs}
\end{document}